%% file: main.tex
\documentclass[10pt,twocolumn,letterpaper]{article}

\usepackage[pagenumbers]{iccv} %

\usepackage{tikz}
\usetikzlibrary{arrows,shapes,calc,matrix,fit,backgrounds}
\usepackage{pgfplots}
\usepackage{pgfplotstable}
\pgfplotsset{compat=1.9}
\usepgfplotslibrary{fillbetween}
\usetikzlibrary{patterns}
\usepackage{multirow}
\usepackage{lipsum}
\usepackage[accsupp]{axessibility}  %

\input{preamble}
\definecolor{iccvblue}{rgb}{0.21,0.49,0.74}
\usepackage[pagebackref,breaklinks,colorlinks,allcolors=iccvblue]{hyperref}
\usepackage{tabularx}
\usepackage{bm}
\usepackage[most]{tcolorbox}
\colorlet{LightLavender}{Lavender!40!}
\colorlet{LightBlue}{SkyBlue!40!}
\colorlet{LightGreen}{LimeGreen!30!}
\definecolor{steelblue}{rgb}{0.27, 0.51, 0.71}
\definecolor{lightsteelblue}{rgb}{0.865, 0.93375, 0.9725}
\definecolor{tomato}{rgb}{0.99, 0.39, 0.28}
\definecolor{lighttomato}{rgb}{0.9975, 0.8475, 0.82}
\usepackage{pgfplots}
\usepgfplotslibrary{polar}
\definecolor{oursrow}{HTML}{E6F2F4}
\usepackage{caption}

\tcbset{on line, boxsep=2pt, left=0pt, right=0pt, top=0pt, bottom=0pt,
        colframe=white, colback=LightLavender, 
        highlight math style={enhanced}
        }

\tcbset{lavenderbox/.style={colback=LightLavender}}
\tcbset{bluebox/.style={colback=LightBlue}}
\tcbset{greenbox/.style={colback=LightGreen}}

\newcommand{\pretraining}{\textsc{\tcbox[lavenderbox]{pretraining}}}
\newcommand{\training}{\textsc{\tcbox[bluebox]{training}}}
\newcommand{\benchmark}{\textsc{\tcbox[greenbox]{benchmark}}}

\def\paperID{3330} %
\def\confName{ICCV}
\def\confYear{2025}

\title{Data Leakage in Visual Datasets}

\author{
Patrick Ramos$^{*}$ \quad Ryan Ramos$^{*}$ \quad Noa Garcia\\
The University of Osaka\\
\tt\small \{{patrickramos@is., ryanramos@is., noagarcia@}\}ids.osaka-u.ac.jp
}

\begin{document}

\maketitle

\def\thefootnote{*}\footnotetext{Equal contribution.}
\renewcommand{\thefootnote}{\arabic{footnote}}

\input{sec/0_abstract}    
\input{sec/1_intro}
\input{sec/2_relatedwork}
\input{sec/3_definition}
\input{sec/4_method}

\input{sec/5_settings_datasets}

\input{sec/6_features_results}
\input{sec/7_leakage_results}

\input{sec/8_conclusion}

\section*{Acknowledgments}
This work was supported by JSPS KAKENHI No. JP23H00497 and JP22K12091.

{
    \small
    \bibliographystyle{ieeenat_fullname}
    \bibliography{main}
}

\appendix
\input{sec/X_suppl}

\end{document}

%% file: sec/0_abstract.tex
 \begin{abstract}
We analyze data leakage in visual datasets. Data leakage refers to images in evaluation benchmarks that have been seen during training, compromising fair model evaluation. Given that large-scale datasets are often sourced from the internet, where many computer vision benchmarks are publicly available, our efforts are focused into identifying and studying this phenomenon. We characterize visual leakage into different types according to its modality, coverage, and degree. By applying image retrieval techniques, we unequivocally show that all the analyzed datasets present some form of leakage, and that all types of leakage, from severe instances to more subtle cases, compromise the reliability of model evaluation in downstream tasks. 
\end{abstract}

%% file: sec/1_intro.tex
\section{Introduction}
\label{sec:intro}

Visual datasets are at the core of advancements in computer vision, serving not only as the foremost resources for model training but also as benchmarks for evaluating technological progress. Datasets have been central to the key milestones in the field, such as ImageNet \cite{deng2009imagenet} for image recognition, COCO \cite{lin2014microsoft} for object detection, and LAION~\cite{schuhmann2021laion} for image generation. Despite their impact, best practices for the creation and use of visual datasets have received little attention. As the demand for larger datasets keeps growing, analyzing their content has become a major challenge. In this context, audits have been crucial for revealing critical issues such as bias~\cite{garcia2023uncurated,meister2023gender}, toxicity~\cite{birhane2021large,birhane2021multimodal,birhane2024into}, or  duplication~\cite{touvron2023llama}, highlighting the urgent need for thorough dataset analysis. 

In this paper, we explore an often-overlooked issue in large visual datasets: data leakage. Data leakage occurs when a model has access to some or all of the evaluation test data during training. This may result in inflated performance metrics and compromise the integrity of model evaluation --- one of the fundamental principles in machine learning. In the era of datasets scraped from the internet \cite{schuhmann2021laion,schuhmann2022laion,gadre2024datacomp}, where most benchmarks for model evaluation are also publicly available online, the problem of data leakage stands out as particularly relevant. A potential consequence is that as large vision and language models (VLMs) \cite{radford2021learning,cherti2023reproducible,li2024llava} are trained on huge pre-training datasets from the web, any existing leakage can result in overly optimistic evaluations, making comparisons to models trained on datasets without leakage unfair. 

Data leakage is an active research topic in the context of large language models (LLMs) \cite{openai2024gpt4technicalreport,touvron2023llama,jiang2023mistral,anil2023palm}, as LLMs are trained on vast amounts of text from the internet, often including partial or complete portions of  evaluation benchmarks \cite{ravaut2024much}. While the natural language processing (NLP) community is actively working to detect and address data leakage~\cite{magar2022data,dodge2021documenting,elazar2024what,piktus-etal-2023-roots,sainz-etal-2023-nlp,golchintime,deng2024investigating,roberts2023cutoff,shi2024detecting,oren2024proving,balloccu-etal-2024-leak},\footnote{There was a dedicated workshop on the topic at ACL 2024: the 1st Workshop on Data Contamination, \url{https://www.aclweb.org/portal/content/first-workshop-data-contamination}.} there has been comparatively little focus on identifying overlaps of images in visual datasets. To close this gap, we analyze a variety of standard visual datasets and explore whether there are overlaps in test and training images. We do so by categorizing datasets into three types according to their standard use: \pretraining{}, \ie datasets used for training large models,  \training{}, \ie datasets for fine-tunning or training smaller models, and \benchmark{}, \ie datasets for reporting models' performance.

We define data leakage across three dimensions: modality, coverage, and degree. \textit{Modality} refers to the type of data being leaked, such as images with or without annotations. \textit{Coverage} describes the relationship between the training and the evaluation splits, such as from the same or different dataset. Finally, \textit{degree} specifies the level of similarity required for two images to be considered leaked, such as identical or near-identical. Each scenario of data leakage is described by a unique combination of these three dimensions. In all cases, data leakage is detected through image retrieval by extracting image representations and conducting an efficient k-nearest neighbor (knn) search. 

Our method and leakage definitions are validated by experimental results in \cref{sec:exp_robust} and extensive qualitative examples are provided in the supplementary material. Following our approach, we conduct a comprehensive leakage analysis in \cref{sec:leakage-analysis}, covering 20 data splits across 7 popular visual datasets. We study both intra-dataset leakage, \ie, image overlap between the test and training splits of the same dataset, and inter-dataset leakage, \ie, image overlap between two different datasets. Additionally, we demonstrate the impact that leaked samples have on the evaluation of three downstream tasks: zero-shot image classification, supervised image classification, and image-to-text retrieval. Our main findings are:
\begin{itemize}
    \item For all the datasets under analysis, data leakage is present within splits of the same dataset, \ie training and tests splits of the same dataset have shared images. 
    \item For all \benchmark{} under analysis, we identify instances of data leakage in the \pretraining{} datasets.
    \item \textit{Hard} leakage rates (\ie identical images) tend to be below $3\%$, whereas \textit{soft} leakage (\ie near-identical images) reaches rates of $7\%$. In total, we find up to $10\%$ leakage.
    \item Both hard and soft leakage have a pronounced effect on the evaluation of downstream tasks. While this generally results in inflated performance metrics, we also observe instances of decreased performance when labels differ between training and testing.
\end{itemize}

Our results consistently show that models are capable of memorizing not only identical leaked samples but also near-identical ones. This poses a great risk to fair model evaluation, especially in the context of zero-shot tasks where models have been trained on different datasets with potentially different leakage rates.

%% file: sec/2_relatedwork.tex
\section{Related work}
\label{sec:relatedwork}

Data leakage in visual datasets is often referred to as image duplication.\footnote{\textit{Duplication} can refer to the presence of identical images within the same dataset split, such as multiple copies of the same image in the training set. This scenario typically does not pose a significant problem. Therefore, we use the term \textit{leakage} to specifically describe instances where duplicates occur between a training and a test set.} Several models that rely on vast amounts of internet-sourced images for training have adopted different techniques to detect leakage between their training and test sets \cite{mahajan2018exploring,radford2021learning,zhai2022lit,cherti2023reproducible}. For instance, using RMAC features \cite{tolias2016particular}, Mahajan \etal \cite{mahajan2018exploring} reported less than $1\%$ overlap between their dataset collected from Instagram\footnote{\url{https://www.instagram.com/}} and four standard validation sets: ImageNet~\cite{deng2009imagenet}, CUB~\cite{wah2011caltech}, Places~\cite{zhou2017places}, and COCO~\cite{lin2014microsoft}. On the other hand, in CLIP \cite{radford2021learning}, Radford \etal trained a custom duplication detector and found substantial overlap with rates reaching up to $21.5\%$ in all but synthetic datasets (MNIST \cite{lecun1998gradient}, CLEVR \cite{johnson2017clevr}) and those created post-training cutoff (ObjectNet \cite{barbu2019objectnet}, Hateful Memes \cite{kiela2020hateful}). Alternatively, in OpenCLIP \cite{cherti2023reproducible}, pHash \cite{zauner2010implementation} was used to estimate the leakage between LAION-400M \cite{schuhmann2021laion} and the downstream datasets, ranging from $1\%$ to $5\%$. 

Leakage in visual datasets has been primarily conducted by the authors of models themselves, often concluding that there is a minimal impact on the evaluations. However, as large datasets and models are later applied to a wide range of tasks beyond their original scope, we believe an independent and systematic assessment is essential.

%% file: sec/3_definition.tex
\section{Data leakage definition}
\label{sec:definition}

Data leakage compromises the integrity of a benchmark by using information from the evaluation split during the training process of a model. This can occur in different forms. To systematically characterize the issue, we break it down into three dimensions: \textit{modality}, \textit{coverage}, and \textit{degree}. %

\vspace{-14pt}
\paragraph{Leakage modality}
is the dimension that defines the type of data being leaked. In visual datasets, we distinguish between two modalities:

\begin{itemize}

     \item \textbf{Image-only leakage}: only the images from the evaluation set are exposed during training, while their corresponding annotations or labels remain unseen.
    
    \item \textbf{Full leakage}: both the images and their associated annotations or labels are  exposed during training.
\end{itemize}

\vspace{-14pt}
\paragraph{Leakage coverage}
is the dimension that defines the relationship between the training and evaluation splits of the leaked samples. We consider two scenarios:

\begin{itemize}

    \item \textbf{Intra-dataset leakage}: occurs when there is an overlap between training and evaluation samples within the same dataset. This type of leakage compromises the integrity of the evaluation protocol for that specific dataset.
    
    \item \textbf{Inter-dataset leakage}: occurs when samples from an evaluation dataset are leaked into a different dataset used in the training process of a model. This type of leakage can affect the generalizability of the model across datasets, and it can be particularly problematic when benchmarking models trained on different datasets.
\end{itemize}

\vspace{-14pt}
\paragraph{Leakage degree} 
is the dimension that defines the level of similarity required between two images for them to be considered leaked. We distinguish between two cases:

\begin{itemize}

    \item \textbf{Hard leakage}: occurs when identical images appear in both the training and evaluation sets. This represents a direct and unambiguous form of leakage, where the overlap between datasets is explicit and easily detectable.
    
    \item \textbf{Soft leakage}: occurs when nearly identical images are present in both the training and evaluation sets. Unlike hard leakage, soft leakage involves images with minor variations. Although soft leakage has been often overlooked, we show in \cref{sec:downstream} that it can greatly impact the evaluation of model generalization.
\end{itemize}

Each leakage scenario is defined by a unique combination of attributes across the three dimensions. For example, an image-only, intra-dataset, and hard leakage scenario describes a situation where the leakage occurs exclusively through exact images within the same dataset.

%% file: sec/4_method.tex
\section{Data leakage detection}
\label{sec:method}
To detect data leakage in visual datasets, we need to identify whether images from a test split have been leaked into a train split. We formulate the problem as an image retrieval task (\cref{sec:image-detection}) and categorize leakage degree into hard and soft  by thresholding (\cref{sec:degree}).

\subsection{Image retrieval}
\label{sec:image-detection}
Given two datasets, one commonly used for training ${\mathcal{T} = \{\bm{x}_o\}}$ and the other for evaluation $\mathcal{E} = \{\bm{x}_q\}$, where $\bm{x}_o$ and $\bm{x}_q$ are images, the goal is to find how many images from $\mathcal{E}$ can be found in $\mathcal{T}$. Using image retrieval terminology, we treat images in $\mathcal{E}$ as \textit{queries} and search for them within $\mathcal{T}$, referred to as the \textit{collection}. 

Each image $\bm{x}$, whether in the query or in the collection, is represented by features obtained from an image encoder $e(\cdot)$, yielding the encoded representations $\bm{c} = e(\bm{x}_o)$ and $\bm{q} = e(\bm{x}_q)$. These image representations must contain enough information for detecting leaked images while being computationally efficient to handle large-scale data and robust enough to account for small transformations between datasets, such as resizing and cropping. Given the size of web-scale datasets, directly extracting representations for every image is computationally expensive. When available, we use pre-computed image representations provided by dataset authors, typically pre-trained CLIP  \cite{radford2021learning}. 

Depending on the size of the collection $|\mathcal{T}|$, image retrieval is conducted as:
\begin{itemize}
    \item \textbf{direct search}: if $|\mathcal{T}|$ is sufficiently small, we match each query representation $\bm{q}$ with each representation in the collection $\bm{c}$ to obtain a similarity score per pair ${s_{q,c} = \text{cos}(\bm{q}, \bm{c})}$, where $\text{cos}$ is cosine similarity;
    \item \textbf{knn search}: for larger datasets, we conduct a knn search with Faiss~\cite{douze2024faiss} by building an index $I$ with the representations in $\mathcal{T}$ and use it for fast search given $\bm{q}$. 
    The search is conducted based on similarities $s_{q,c}=\text{faiss}_{\text{sim}}(\bm{q}, \bm{c}, \bm{I})$, where $\text{faiss}_{\text{sim}}$ returns the cosine similarity between $\bm{q}$ and the indexed representation of $\bm{c}$ in $I$.
\end{itemize}

\subsection{Leakage degree}
\label{sec:degree}

The leakage degree is the dimension that defines how similar two images need to be to classify them as leakage. We use the similarity score \( s_{q,c} \) obtained through direct or knn search to identify whether a pair of images are identifcal (hard leakage) or near-identical (soft leakage).

For a given pair of images, they are considered hard leakage when the similarity score is above a threshold
\begin{equation}
    s_{q,c} \geq \tau_h ,
\end{equation}
whereas if the similarity score falls within 
\begin{equation}
    \tau_s \leq s_{q,c} < \tau_h
\end{equation}
the pair is considered soft leakage. The hard and soft leakage rates, $\text{H}$ and $\text{S}$ respectively, are computed as
\begin{equation}
    \text{H} = \frac{N_h}{|\mathcal{E}|}, \qquad \text{S} = \frac{N_s}{|\mathcal{E}|},
\end{equation}
where $N_h$ and $N_s$ are the number of hard and soft leakage samples found, respectively. The thresholds $\tau_s$ and $\tau_h$ are found empirically and validated in \cref{sec:threshold-val}.

\input{tables/datasets}

%% file: tables/datasets.tex
\begin{table}[t]
\centering
\caption{Datasets details by type.}

\vspace{-3pt}
\resizebox{\columnwidth}{!}{
\begin{tabular}{@{}l r r l l}
\toprule
dataset & split & images & source & type\\
\midrule 

coco & test2014 & $40k$ & flickr & \footnotesize \benchmark{} \\ 
coco & test2015 & $81k$ & flickr & \footnotesize \benchmark{} \\ 
coco & test2017 & $40k$ & flickr & \footnotesize \benchmark{} \\ 
flickr30k & all & $31k$ & flickr & \footnotesize  \benchmark{} \\
gcc & val & $13k$ & flume & \footnotesize \benchmark{} \\
imagenet & test & $100k$ & flickr & \footnotesize \benchmark{} \\
imagenet & val & $50k$ & flickr & \footnotesize \benchmark{} \\
openimages & test & $125k$ & flickr & \footnotesize \benchmark{} \\
openimages & val & $41k$ & flickr & \footnotesize \benchmark{} \\
textcaps & test & $3k$ & openimages & \footnotesize \benchmark{} \\

coco & val2014 &  $40k$ & flickr & \footnotesize \benchmark{} \training{} \\
coco & val2017 & $5k$ & flickr & \footnotesize \benchmark{} \training{} \\
coco & train2014 & $82k$ & flickr & \footnotesize \training{}\\
coco & train2017 & $118k$ & flickr & \footnotesize \training{}\\
coco & unlabeled & $123k$ & flickr & \footnotesize \training{}\\
textcaps & train & $25k$ & openimages &  \footnotesize \training{}\\
gcc & train & $2,874k$ & flume & \footnotesize \training{} \pretraining{}\\
imagenet & train & $1,281k$ & internet & \footnotesize \training{} \pretraining{} \\
openimages & train & $1,743k$ & flickr & \footnotesize \training{} \pretraining{}\\
laion & all &$407,314k$ & commoncrawl & \footnotesize \pretraining{} \\

\bottomrule
\end{tabular}
}
\label{tab:datasets}
\end{table}

%% file: sec/5_settings_datasets.tex
\section{Experimental settings}

\paragraph{Setup}
Our default image encoder, $e(\cdot)$, is a pretrained CLIP ViT-B/32 \cite{radford2021learning}. %
In the knn search, we use the \mbox{AutoFaiss}\footnote{\url{https://github.com/criteo/autofaiss}} implementation for Faiss. %

\vspace{-12pt}
\paragraph{Datasets}
We categorize datasets into three types. \pretraining{} are large datasets sourced from the Internet and used to train large models. \training{} are standard sets, typically annotated, used for model training or finetuning. \benchmark{} are smaller, annotated datasets used for evaluation. As summarized in \cref{tab:datasets}, our analysis comprises 20 data splits across 7 datasets:

\begin{description}[noitemsep,nolistsep]
    \item[COCO]~\cite{lin2014microsoft} contains several splits (train, val, test, unlabeled) across different versions (2014, 2015, 2017). The images were collected from Flickr\footnote{\url{https://www.flickr.com/}}, and each image includes object annotations and five captions. COCO is the standard dataset for object detection \cite{ren2016faster,wang2025yolov9} and image captioning \cite{vinyals2015show}. The dataset is used in different ways in the literature. While some papers \cite{girshick2015fast,liu2021swin,alayrac2022flamingo} report results on the test splits, which require evaluation via a server, others \cite{carion2020end} use the validation split, and some \cite{ren2016faster} report results on both. Additionally, it is common to augment the training data with the validation set to improve performance \cite{karpathy2015deep}. %
    We use the test and val splits as \benchmark{}, with the val also serving as \training{}, along with the train and unlabeled splits. 
    \item[Flickr30k]~\cite{young-etal-2014-image} has a single split with $31,783$ images collected from Flickr, where each image is annotated with a caption. Train, val, and test splits were later introduced in \cite{karpathy2015deep}. %
    We use it as \benchmark{}. %
    \item[GCC]~\cite{sharma2018conceptual} contains about $3.3$ million images from the internet collected through the crawler Flume \cite{chambers2010flumejava}. Images are divided into train and val splits, and each image is paired with an alt-text caption. Due to broken links, we could only download $2,874,229$ train and $13,354$ val images. The val split is used to report results \cite{chen2024would}, making it a \benchmark{}. Meanwhile, the training split has been used for training VLMs \cite{Su2020VL-BERT,10.1007/978-3-030-58577-8_7}, so we use it as \pretraining{}.
    \item[ImageNet]~\cite{deng2009imagenet} has more than a million images divided into train, val, and test splits. Images are collected from the internet\footnote{While the val and test images are from Flickr, the source of the train images is not specified.} and annotated with class labels. Given its widespread use for training visual models that serve as backbones for downstream tasks \cite{he2016deep,caron2021emerging}, we use its training split as \pretraining{}. The test and val splits are used as \benchmark{}.
    \item[LAION]~\cite{schuhmann2021laion} contains about $400$ million images crawled with Common Crawl\footnote{\url{https://commoncrawl.org/}} from the internet. Each image is paired with an alt-text caption. LAION is designed for training large VLMs \cite{rombach2021highresolution,cherti2023reproducible}, and does not provide data splits. We use the entire dataset as \pretraining{}.
    \item[OpenImages]~\cite{kuznetsova2020open} has several versions. We use OpenImages v4, which contains about $2$ million images from Flickr split into train, val, and test. The dataset is commonly used for image classification and object detection. We use the val and test splits as \benchmark{} and the training split as \pretraining{}.
    \item[TextCaps]~\cite{sidorov2020textcaps} is a subset of OpenImages with human annotated captions. Data is split into $3,353$ test and $25,119$ training images, which we use as \benchmark{} and \training{}, respectively.
\end{description}

%% file: sec/6_features_results.tex
\section{Method evaluation}
\label{sec:exp_robust}
We validate the proposed method by showing the efficacy of the image retrieval (\cref{sec:image-detection-val}) and inspecting the choice of thresholds in the leakage degree (\cref{sec:threshold-val}).

\subsection{Image retrieval evaluation}
\label{sec:image-detection-val}
To be able to detect small variations in images such as cropping or scaling, we evaluate the data leakage detection method with a particular focus on the cases where images undergo non-semantic transformations. We conduct this evaluation on Flickr30k. We use the full dataset, \ie, $31,783$ images, as the collection, and we randomly choose $5,000$ images as queries. Given a query image (with or without a non-semantic transformation), the goal is to retrieve the original image from the collection. 

The evaluation flow is as follows: for images in the collection, we just extract their encoded representations and store them. For query images, we apply a non-semantic transformation before feeding them to the image encoder. Then, the encoded representation of a transformed query is matched against all the representations in the collection with direct search, and the collection image with the highest cosine similarity is retrieved. Accuracy is measured as recall at 1 (R@1), \ie, the number of retrieved images corresponding to their original query over the number of queries.

\vspace{-10pt}
\paragraph{Image encoders} We evaluate three types of image encoders: \emph{resnet}, \emph{dino2}, and \emph{clip}. Image representations from \emph{resnet} are extracted from the second-to-last layer of a 152-layer ResNet \cite{he2016deep} pre-trained on ImageNet. The \emph{dino2} encoder is a DINOv2 ViT-B/14 \cite{oquab2023dinov2}, while \emph{clip} is the pre-trained CLIP ViT-B/32 \cite{radford2021learning} image encoder. 

\vspace{-10pt}
\paragraph{Image transformations} We use four types of non-semantic transformations:

\begin{itemize}
    \item \textbf{Geometric}: vertical (\textit{flip-v}) and horizontal  (\textit{flip-h}) flips, and $D$ degrees rotations with $D = \{45, 135, 225, 315\}$, namely \textit{rot-45}, \textit{rot-135}, \textit{rot-225}, and \textit{rot-315}. 
    \item \textbf{Cropping}: removes $B$ pixels from all four sides of the image, with $B = \{20, 50, 100\}$, namely \textit{crop-20}, \textit{crop-50}, and \textit{crop-100}. 
    \item \textbf{Pixelization}: Gaussian filters (\textit{gauss}), Gaussian noise (\textit{noise}), and downsizing the image to $S$ pixels, with $S = \{128, 256\}$ as \textit{rs-128}, \textit{rs-256}. 
    \item \textbf{Color}: grayscale (\textit{gray}), color inversion (\textit{invert}), and red, green, and blue colorizations (\textit{red}, \textit{green}, \textit{blue}).
\end{itemize}

Examples of the transformations can be found in the supplementary material.

\input{figures/radar-chart}

\vspace{-10pt}
\paragraph{Results} Results are shown in \cref{fig:radar}, where each transformation is plotted along the circular axis, with the lines representing R@1 for the different image encoders. When query images are not transformed (\textit{original}), the three encoders achieve a 100\% performance. For transformed images, the \textit{clip} encoder performs well in cropping, horizontal flipping, noise, and resizing, and it outperforms \textit{resnet} in colorization and pixelation. The most robust encoder for colorization is \textit{dino2}, probably due to its self-supervised training process. However, \textit{dino2} underforms at geometric and cropping transformations, which are the most common type of transformation observed when images are duplicated across datasets. Given this and considering that the largest dataset in our analysis (\ie, LAION) provides pre-computed clip embeddings, \textit{clip} stands out as our preferred choice, offering the best balance between robustness to transformations and computational efficiency.

\subsection{Threshold choice}
\label{sec:threshold-val}

Next, we examine the definition of \textit{hard} and \textit{soft} leakage together with the choice of thresholds. Using the same experimental set-up as in \cref{sec:image-detection-val}, we compute the true positive rate (TPR) and false positive rate (FPR) for different thresholds under two conditions: when query images are not transformed (\textit{original}) and when query images are transformed as in \cref{sec:image-detection-val} (\textit{trans}). The receiver operating characteristic (ROC) curves are plotted in \cref{fig:roc}. For the original case, the area under the curve (AUC) is near perfect,\footnote{Computed AUC value of $0.99999999370713$.} while when query images suffer from some non-semantic transformation, the AUC is still remarkably high ($0.98$).

With respect to the leakage degree, we inspect the choice of hard and soft thresholds in detail:  

\begin{itemize}
    \item \textit{Hard leakage} is defined as the leakage of identical images. We choose $\tau_h = 0.98$ as hard leakage threshold, which achieves a FPR of $0.0$, and a TPR of $1.00$ (original) and $0.08$ (trans). This aligns precisely with the strict definition of hard leakage for identical images: no false positives occur, and true positives are detected only when the image is either identical (original) or undergoes nearly imperceptible transformations.

    \item \textit{Soft leakage} is defined as the leakage of near-identical images. We choose $\tau_s = 0.95$ as soft leakage threshold, which results in a higher TPR of $1.00$ (original) and $0.16$ (trans), while the FPR remains extremely low at a maxiumum of $2.08 \times 10^{-7}$. Again, this is  the expected behavior for \textit{soft} leakage detection, where images that undergo some transformations and hence, are not exactly identical, can be detected while maintaining a very low false positive rate.

\end{itemize}

Our choice of thresholds prioritizes a low FPR over a high TPR to ensure that any detected image is truly a leaked image. As a result, our analysis and findings are conservative: there may be additional leaked images that are not detected. A qualitative analysis of the threshold selection is provided in the supplementary material.

\input{figures/auc-roc}

%% file: figures/radar-chart.tex
\begin{figure}
\centering
\hspace{-25pt}
\begin{tikzpicture}
\begin{polaraxis}[
    width=.75\linewidth,
    xtick={0,18.95,37.9,56.95,75.8,95.75,113.7,132.65,151.6,170.55,189.5,208.45,227.4,246.35,265.3,284.25,303.2,322.15,341.1,360},
    xticklabels={original,flip-v,flip-h,rot-45,rot-135,rot-225,rot-315,crop-20,crop-50,crop-100,gauss,noise,rs-128,rs-256,gray,invert,red,green,blue},
    ytick={0,0.20,0.40,0.60,0.80,1},
    ylabel near ticks,
    ylabel shift = -2pt,
    xticklabel style={font=\footnotesize},
    ytick=\empty,
    ymin=0,
    ymax=1,
    grid=both,                    %
    minor y tick num=1,           %
    grid style={dashed,gray!30},  %
    major grid style={thick},     %
    legend cell align={left},
    legend style={at={(1.3,1.1)},anchor=north,font=\footnotesize}
]
        \addplot[thick, fill=blue, fill opacity=0.2, color=blue] coordinates {
        (0,1)
        (18.95,0.661)
        (37.9,1)
        (56.95,0.8692)
        (75.8,0.4302)
        (95.75,0.4254)
        (113.7,0.8732)
        (132.65,1)
        (151.6,0.9906)
        (170.55,0.6794)
        (189.5,0.357)
        (208.45,0.6314)
        (227.4,0.9944)
        (246.35,1)
        (265.3,0.9922)
        (284.25,0.4024)
        (303.2,0.559)
        (322.15,0.6788)
        (341.1,0.4036)
        (360,1) %
        };
        \addlegendentry{resnet}

        \addplot[thick, fill=yellow, fill opacity=0.2, color=yellow] coordinates {
        (0,1)
        (18.95,0.419)
        (37.9,0.8122)
        (56.95,0.3276)
        (75.8,0.1886)
        (95.75,0.1418)
        (113.7,0.4476)
        (132.65,0.8634)
        (151.6,0.7564)
        (170.55,0.488)
        (189.5,0.6748)
        (208.45,0.8484)
        (227.4,0.958)
        (246.35,0.9976)
        (265.3,0.9854)
        (284.25,0.8522)
        (303.2,0.9164)
        (322.15,0.954)
        (341.1,0.9452)
        (360,1) %
        };
        \addlegendentry{dino2}

        \addplot[thick, fill=red, fill opacity=0.2, color=red] coordinates {
        (0,1)
        (18.95,0.5472)
        (37.9,0.9996)
        (56.95,0.8968)
        (75.8,0.345)
        (95.75,0.3496)
        (113.7,0.89)
        (132.65,0.9994)
        (151.6,0.9846)
        (170.55,0.6376)
        (189.5,0.3856)
        (208.45,0.935)
        (227.4,0.9966)
        (246.35,0.9996)
        (265.3,0.9574)
        (284.25,0.1714)
        (303.2,0.695)
        (322.15,0.755)
        (341.1,0.7906)
        (360,1) %
        };
        \addlegendentry{clip}
\end{polaraxis}
\end{tikzpicture}
\caption{Leakage detection R@1 on $5,000$ query images from the Flickr30k dataset under several transformations.} 
\label{fig:radar}
\end{figure}
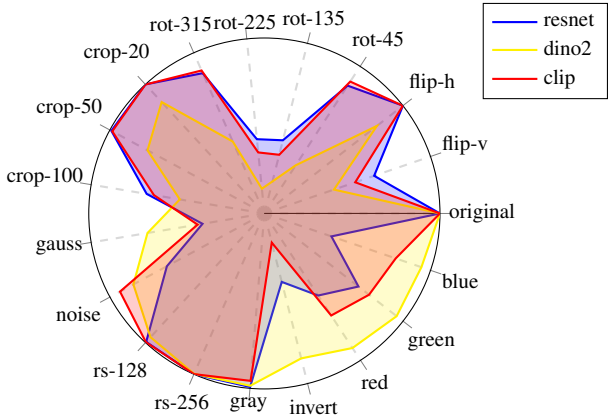

%% file: figures/auc-roc.tex
\begin{figure}
\centering
\hspace{-25pt}
\begin{tikzpicture}
    \begin{axis}[
        width=.71\linewidth,
        xlabel={FPR},
        ylabel={TPR},
        xmin=-0.01, xmax=1,
        ymin=0, ymax=1.01,
        xtick={0,0.2,0.4,0.6,0.8,1},
        ytick={0,0.2,0.4,0.6,0.8,1},
        xticklabel style={font=\footnotesize},
        yticklabel style={font=\footnotesize},
        legend pos=south east,
        legend cell align={left},
        legend style={font=\footnotesize},
        grid=major,
        grid style={dashed, gray!30}]

        \addplot[ultra thick,
            color=orange,
            mark=none,
            smooth
        ] coordinates {
            (0.9999999937071299, 1.0)
            (0.9999990560694734, 1.0)
            (0.9999762381222076, 1.0)
            (0.9997040589012649, 1.0)
            (0.9976402429047888, 1.0)
            (0.9869663897803789, 1.0)
            (0.9489074004153294, 1.0)
            (0.8520131017557108, 1.0)
            (0.6731427285885092, 1.0)
            (0.43818187653388707, 1.0)
            (0.225713322006167, 1.0)
            (0.09270093134478635, 1.0)
            (0.0319065194135045, 1.0)
            (0.009794984582468063, 1.0)
            (0.002732804732238374, 1.0)
            (0.0006447611855767416, 1.0)
            (0.00011230256119816248, 1.0)
            (1.0087470895475426e-05, 1.0)
            (5.7453904725945505e-06, 1.0)
            (3.1023849977974953e-06, 1.0)
            (1.5165817129192624e-06, 1.0)
            (5.663583160279403e-07, 1.0)
            (2.0766471587691145e-07, 1.0)
            (1.8878610534264676e-08, 1.0)
            (0.0, 1.0)
            (0.0, 1.0)
            (0.0, 1.0)
            (0.0, 0.0)
        };
        \addlegendentry{original (AUC = 1.00)}

        \addplot[ultra thick,
            color=teal,
            mark=none,
            dashed
        ] coordinates {
            (0.9999999509818533, 1.0)
            (0.9999987715655007, 1.0)
            (0.9999812853352944, 1.0)
            (0.9998019040900342, 1.0)
            (0.998522931218929, 1.0)
            (0.9920299755902874, 0.9999894736842105)
            (0.9677072729019073, 0.9999473684210528)
            (0.8991824087782226, 0.9997263157894736)
            (0.7543598660612263, 0.9986315789473684)
            (0.5303546575519409, 0.9942947368421052)
            (0.2907889179244127, 0.9826842105263157)
            (0.1200275127596223, 0.9607368421052632)
            (0.037673751444876113, 0.9285473684210528)
            (0.009408312219097867, 0.8806947368421053)
            (0.00197210602492639, 0.8056631578947369)
            (0.00034790132779560757, 0.6821789473684211)
            (4.518578871191572e-05, 0.5021789473684212)
            (2.908299633357512e-06, 0.31349473684210527)
            (1.4741876401405626e-06, 0.2807684210526316)
            (6.991709971549602e-07, 0.250021052631579)
            (2.775486952230492e-07, 0.21932631578947365)
            (9.373064528415621e-08, 0.1892105263157895)
            (2.6827499180270855e-08, 0.16094736842105262)
            (4.636851710170272e-09, 0.13535789473684212)
            (0.0, 0.11167368421052631)
            (0.0, 0.08041052631578947)
            (0.0, 0.021273684210526316)
            (0.0, 0.0)
        };
        \addlegendentry{trans (AUC = 0.98)}

    \end{axis}
\end{tikzpicture}
 
\caption{ROC curve across different thresholds on the Flickr30k dataset. For images without  transformations (original), the retrieval achieves near-perfect performance with an AUC of almost $1.00$, and even when images undergo transformations (trans), the performance remains high, with an AUC of $0.98$.} 
\label{fig:roc}
\end{figure}
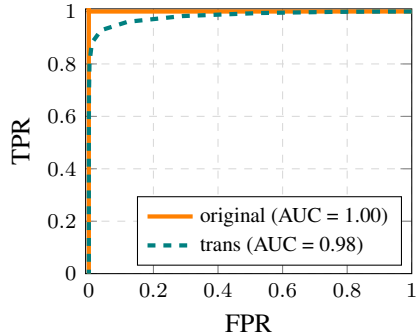

%% file: sec/7_leakage_results.tex
\section{Leakage analysis}
\label{sec:leakage-analysis}
Focusing on the image-only modality, we corroborate the existence of intra-dataset (\cref{sec:intra}) and inter-dataset (\cref{sec:inter}) leakage on several widely-used benchmarks. After that, we present a comprehensive analysis on the impact of leakage on the evaluation of downstream tasks, including how leakage modality and degree affects the overall performance of benchmarks (\cref{sec:downstream}).

\input{figures/inter-leakage}

\subsection{Intra-dataset leakage}
\label{sec:intra}

To compute intra-dataset leakage (\ie, image overlap between training and evaluation samples within the same dataset) we use datasets with both \benchmark{} and \training{}/\pretraining{} splits. That leaves us with five datasets: COCO, GCC, ImageNet, OpenImages, and TextCaps. For COCO, test splits are compared against train, val, and unlabeled splits, and the val splits against the train and unlabeled splits. We note that the val2014 split is included in train2017.

\input{figures/intra-leakage}

\vspace{-10pt}
\paragraph{Results} 
Results are shown in \cref{fig:intradataset}. 
ImageNet test and val, GCC val, and COCO val 2017 are the most leaked datasets. The ImageNet test split has a hard leakage of $1.54\%$ and a soft leakage of $1.95\%$, while the val split has a slightly higher hard leakage of $1.58\%$ but a lower soft leakage of $1.78\%$. %
The second highest leaked dataset is GCC followed by COCO, in which the val2017 split has a soft leakage of $3\%$. COCO test splits also present a soft intra-dataset leakage between $1.35\%$ and $1.38\%$.
Both the test and val splits of OpenImages have small rates of hard leakage, $0.05\%$, but larger rates of soft leakage, $1.36\%$ and $1.38\%$, respectively. 
TextCaps has the smallest leakage rate, $0.69\%$, and we do not find any hard leakage.

These results show that intra-dataset leakage occurs in \textit{all} the analyzed datasets, either in its hard or soft degree. Given that data leakage can compromise model evaluation and that datasets are specifically designed for this purpose, intra-dataset leakage is a risk that \textit{by design} should not exist. Yet, we have identified multiple instances in all datasets.

\input{figures/leakage-examples-main}

\subsection{Inter-dataset leakage}
\label{sec:inter}
To compute intra-dataset leakage (\ie, leakage between different datasets when a model has been trained on \pretraining{} and evaluated on a \benchmark{}), we use GCC train, ImageNet train, OpenImages train, and LAION as collection data and COCO 2014 test and val, Flickr30K, TextCaps test, OpenImages test and val, and ImageNet test and val as query data.

We extract CLIP ViT-B/32 embeddings in all datasets except LAION, in which we use the provided precomputed embeddings.\footnote{\url{https://laion.ai/blog/laion-400-open-dataset/}} We found that LAION's precomputed embeddings do not fully match embeddings extracted by CLIP's official implementation.\footnote{\url{https://github.com/openai/CLIP}} This is addressed by rescaling query images as in the \texttt{clip-retrieval}\footnote{\url{https://github.com/rom1504/clip-retrieval}} repository.
We conduct a knn search by building a single index $I$ per collection dataset, except for LAION, which we found that due to its size, a single index yielded poor retrieval. Instead, we create an index per data partition, with each partition containing a million images. 

\vspace{-14pt}
\paragraph{Results} Results are shown in \cref{fig:interleakage}, where the columns are \benchmark{} used as queries and the rows are \pretraining{} used as collections. The left part, in red, reports the hard leakage rate, whereas the right part, in blue, is the soft leakage rate, both expressed as a percentage. Overall, leakage exists in all datasets. 

Figure \ref{fig:interleakage_exact} shows that while most of the rates are relatively small, the highest rates of hard leakage are found in LAION, specifically for the TextCaps and OpenImages test splits, with about $3\%$ hard leaked images (\ie, identical images). OpenImages train split contains about $1.5\%$ hard leaked images from COCO test and val splits. ImageNet, which is one of the most common datasets for pretraining, also contains hard leaked images from all the benchmarks. Among the \pretraining{} datasets, GCC has the least hard leakage, with all rates remaining below $1\%$. Soft leakage (\ie, near-identical images) is shown in \cref{fig:interleakage_semantic}. %
The dataset with the most soft leakage is
LAION, ranging from $1.09\%$ to $7.91\%$. 
When comparing different \benchmark{}, OpenImages test and val splits and TextCaps test split are the most leaked ones, while Flickr30k shows the least soft leakage rates. 
Examples of both hard and soft leakage are shown in \cref{fig:examples_main}. While hard-leaked images are identical, soft-leaked samples are also extremely similar, often depicting the same person or object in different poses. 

In total, ImageNet, OpenImages, TextCaps, COCO, and Flickr30k have $4.64\%$, $10.67\%$, $9.60\%$, $4.59\%$, and $1.81\%$ of their test samples leaked into LAION, respectively. Although leakage rates may represent only a small fraction of the \benchmark{} datasets, these results are worrisome, as a model can potentially memorize the features of the leaked test images and affect the downstream evaluation, as we will see in \cref{sec:downstream}.

\subsection{Impact on downstream evaluation}
\label{sec:downstream}
The next natural step is to study the impact of data leakage 
on downstream evaluations. In particular, we  evaluate three tasks: zero-shot  classification, supervised classification, and text-image retrieval. In each task, we evaluate the performance of a pretrained model on different subsets of a \benchmark{} in which we know which samples have been leaked into the \pretraining{} dataset. Then, we compare the results on different subsets of the benchmark: 
\begin{itemize}
    \item \textbf{original dataset}: the whole benchmark containing $N$ samples.
    \item \textbf{leaked set}: a subset of $A$ samples identified as leaked.
    \item \textbf{non-leaked set}: a subset of $N-A$ samples that have not been identified as leaked.
    \item \textbf{random set}: a subset of $A$ samples randomly selected from the original dataset, serving as a control set.
\end{itemize}

\vspace{-10pt}
\paragraph{Zero-shot classification} 
As VLMs tend to not fully disclose their training data \cite{radford2021learning}, it is usually impossible to identify their leaked samples. For this analysis, we rely on OpenCLIP ViT-B-32\footnote{\url{https://github.com/mlfoundations/open_clip}} pretrained on LAION, and evaluate it on the ImageNet val split following \cite{radford2021learning}. 

\input{tables/openclip}

Accuracy results for the different subsets on hard and soft leakage are shown in \cref{tab:openclip}. For both hard and soft leakage, the subset of leaked images consistently achieves much higher accuracy than the non-leaked images ($+14.11$ for soft leakage) and the randomly selected images ($+12.81$ for soft leakage). This strongly suggests that the model benefits from exposure to leaked images during training. Moreover, the performance of the non-leaked subset decreases from $54.15$ to $53.54$ when we exclude not only identical images but also near-identical images. %
This corroborates that near-identical images (\ie soft leakage) are contributing to improved performance when seen during training. In a zero-shot scenario where models are trained on different \pretraining{} datasets, this is particularly problematic, as the number of leaked images may vary between models, making comparisons between them especially unfair.

\vspace{-10pt}
\paragraph{Supervised classification} We check how the intra-dataset leakage on ImageNet affects two standard backbones, ResNet50 and ResNet152, both pretrained on the train split and evaluated on different subsets of the val split.

The results are in \cref{tab:resnet}. At first sight, it seems that hard leakage is greatly detrimental to model performance, reducing accuracy from $80.37\%$ on the original dataset to just $30.16\%$. An in-depth inspection reveals the cause of this behavior: the labels. Unlike in the zero-shot task, where labels are not used during training, for  supervised training, we observe full leakage. We find that 98.61\% of the hard-leaked images have different labels in the train and val splits of ImageNet. In the samples in which the leaked labels are the same, the accuracy raises to $100\%$, while in the samples with different labels, accuracy is reduced to $29.18\%$ and $30.72\%$, demonstrating that both ResNet50 and ResNet152 are able to memorize leaked data. This also occurs in the soft-leaked subset: leaked images where the label is the same get $96.06\%$ and $95.44\%$ accuracy, in contrast to those where it is different, with accuracies of $30.99\%$ and $32.26\%$, respectively.

\input{tables/resnet}

\input{tables/retrieval}

\vspace{-10pt}
\paragraph{Image-to-text retrieval} Other than image classification, we study the impact of data leakage on another task: image-to-text retrieval. Given a query image, the goal is to find its associated caption within a collection of captions. We evaluate OpenCLIP ViT-B-32 trained on LAION on different subsets of the Flickr30k. For the results to be comparable between subsets, we fix the number of queries to $200$ images and the size of the collection to $31,783$ captions. We repeat each experiment 10 times with $200$ random images from each subset and report results as the mean and standard deviation of retrieval at $k$ (R@$k$), $k=1,5,10$. 

Results are shown in \cref{tab:retrieval}. %
The best performance is achieved on the hard-leaked subset, being  $+9.55$ and $+13.05$ better than the non-leaked subset on R@1 and R@10, respectively. Soft leakage has similar behavior, with soft-leaked images being $+4.5$ R@10 over the equivalent non-leaked images. While in this case most subsets outperform on average the full dataset results, this can be attributed to the use of a much smaller query size of $200$ instead of the original $31,783$ images in Flickr30k. Regardless, the non-leaked subsets tend to have a larger variance than the leaked subsets, meaning that the latter produce more stable results, probably due to the model being familiar with the leaked samples seen at training time.

%% file: figures/inter-leakage.tex
\begin{figure*}
\centering
\begin{subfigure}[b]{0.4\linewidth}
\hspace{-70pt}
\centering
\begin{tikzpicture}[scale=0.9]
  \foreach \y [count=\n] in {
      {0.02,0.03,0.03,0.03,0.06,0.05,0.03,0.03},
      {0.80,1.46,1.55,0.03,0.05,0.05,0.15,0.20},
      {0.44,0.54,0.46,0.57,0.32,0.32,1.54,1.58},
      {0.72,0.46,0.56,3.01,2.90,2.83,0.62,0.54},
    } {
    \foreach \j [count=\n] in {Flickr30k,coco-test2014,coco-val2014,textcaps-test, openimages-test,openimages-val,imagenet-test,imagenet-val} {
    \node[rotate=90, anchor=west, font=\footnotesize,scale=0.9] at (\n,-10pt) {\j};
    }
      \foreach \x [count=\m] in \y {
        \node[fill=tomato!\x!lighttomato, minimum size=8mm, text=black, font=\footnotesize,scale=0.92] at (\m,-\n) {\x};
      }
    }

  \foreach \a [count=\i] in {GCC,OpenImages,ImageNet,LAION} {
    \node[minimum size=8mm, font=\footnotesize, anchor=east, scale=0.9] at (10pt,-\i) {\a};
  }
\end{tikzpicture}
\subcaption{hard leakage}
\label{fig:interleakage_exact}
\end{subfigure}
\begin{subfigure}[b]{0.4\linewidth}
\vspace{10pt}
\hspace{-10pt}
\centering
\begin{tikzpicture}[scale=0.9]
  \foreach \y [count=\n] in {
      {0.02,0.33,0.28,0.42,0.71,0.71,0.33,0.30},
      {0.15,0.69,0.63,0.84,1.36,1.38,0.47,0.49},
      {0.06,0.38,0.40,0.45,0.66,0.67,1.95,1.78},
      {1.09,4.13,3.71,6.59,7.77,7.91,4.02,4.02},
    } {
        \foreach \j [count=\n] in {Flickr30k,coco-test2014,coco-val2014,textcaps-test, openimages-test,openimages-val,imagenet-test,imagenet-val} {
    \node[rotate=90, anchor=west, font=\footnotesize,scale=0.9] at (\n,-10pt) {\j};
    }
      \foreach \x [count=\m] in \y {
        \node[fill=steelblue!\x!lightsteelblue, minimum size=8mm, text=black,font=\footnotesize,scale=0.92] at (\m,-\n) {\x};
      }
    }

  \foreach \a [count=\i] in {GCC,OpenImages,ImageNet,LAION} {
    \node[minimum size=8mm, font=\footnotesize, anchor=east, scale=0.9] at (10pt,-\i) {\a};
  }
\end{tikzpicture}
\subcaption{soft leakage}
\label{fig:interleakage_semantic}
\end{subfigure}
\caption{Inter-dataset leakage. Columns indicate \benchmark{} and rows \pretraining{} datasets. Left is for hard leakage (in red), and right for soft leakage (in blue).}
\label{fig:interleakage}
\end{figure*}

%% file: figures/intra-leakage.tex
\definecolor{steelblue}{rgb}{0.27, 0.51, 0.71}
\definecolor{tomato}{rgb}{0.99, 0.39, 0.28}

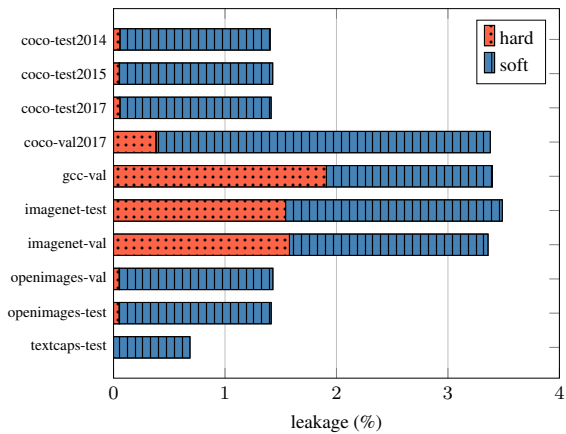
\begin{figure}
\centering
\hspace{-25pt}
\scalebox{0.9}{
\begin{tikzpicture}
\begin{axis}[
    xbar stacked,
    font=\footnotesize,
    width  = 0.98\linewidth,
    xlabel=leakage (\%),
    symbolic y coords={textcaps-test, openimages-test, openimages-val, imagenet-val, imagenet-test, gcc-val, coco-val2017, %
    coco-test2017, coco-test2015, coco-test2014},
    ytick=data,
    x tick label style={rotate=0, font=\footnotesize, xshift=0pt},
    y tick label style={rotate=00, font=\footnotesize, xshift=0pt,scale=0.8},
    xmin=0,xmax=4,
    bar width=9pt,
    legend pos=north east,legend style={legend columns=1, font=\footnotesize},
    grid=none,
    legend cell align={left},
    xmajorgrids=true,ymajorticks=true,yminorticks=false,xminorgrids=true,
    legend image code/.code={\draw [#1] (0cm,-0.1cm) rectangle (0.15cm,0.2cm); },
]
    \addplot[xbar, fill=tomato,  postaction={pattern=dots}] coordinates {
        (0.0564071122,coco-test2014) 
        (0.051575509,coco-test2015) 
        (0.05655274158,coco-test2017)
        (0.38,coco-val2017)
        (1.909540213,gcc-val)
        (1.543,imagenet-test)
        (1.578,imagenet-val) 
        (0.05022481584,openimages-test)
        (0.05045651129,openimages-val)
        (0,textcaps-test)
        };
    \addplot[xbar, fill=steelblue,  postaction={pattern=vertical lines}] coordinates {
        (1.348865727,coco-test2014) 
        (1.376574895,coco-test2015) 
        (1.357265798,coco-test2017) 
        (3,coco-val2017)
        (1.490190205,gcc-val)
        (1.945,imagenet-test) 
        (1.782,imagenet-val)
        (1.364042221,openimages-test) 
        (1.379144642,openimages-val)
        (0.685952878,textcaps-test)
        };
    \legend{hard,soft}
\end{axis}
\end{tikzpicture}
}
\caption{Intra-dataset leakage. Each dataset split in the y-axis is matched against the samples in their corresponding train split.} 
\label{fig:intradataset}
\end{figure}

%% file: figures/leakage-examples-main.tex
\begin{figure*}[tb]
  \centering
\hspace{-25pt}
\begin{subfigure}[b]{0.45\linewidth}
    \centering
    \begin{tabular}{@{\hspace{3pt}}c@{\hspace{3pt}}c@{\hspace{8pt}}c@{\hspace{3pt}}c@{\hspace{3pt}}}

    \includegraphics[width=57pt,height=57pt]{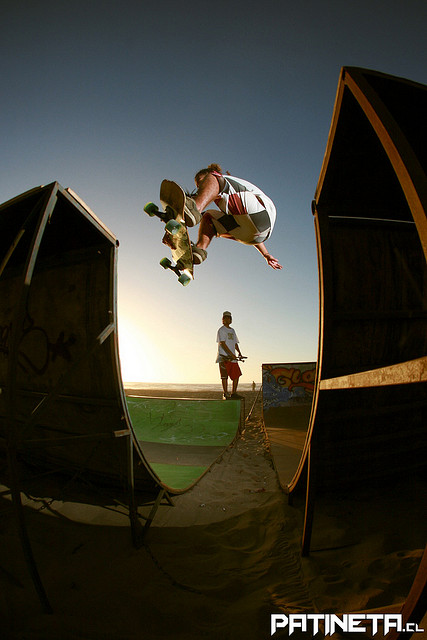} & \includegraphics[width=57pt,height=57pt]{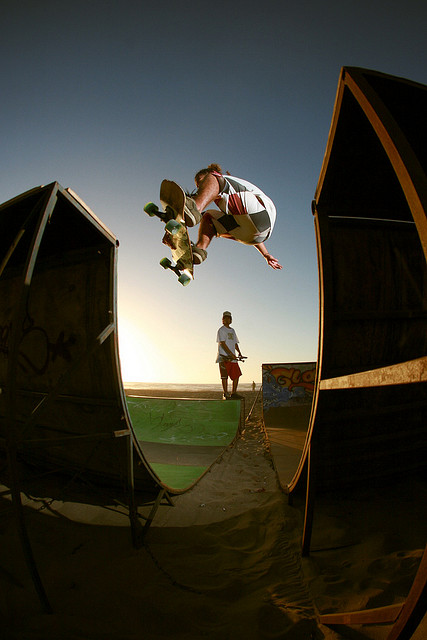} & \includegraphics[width=57pt,height=57pt]{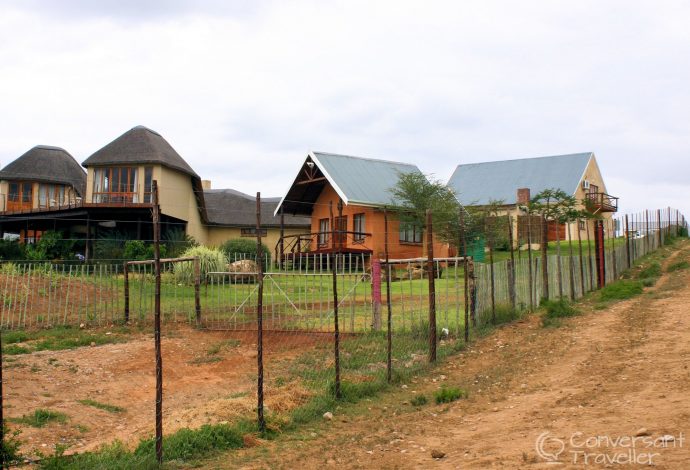} & \includegraphics[width=57pt,height=57pt]{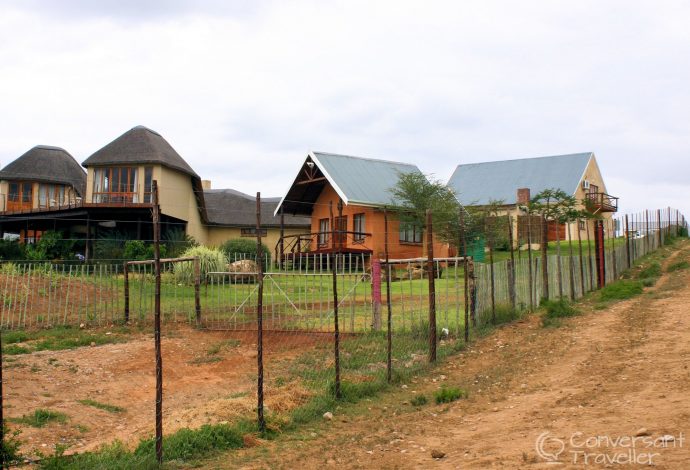}\\[10pt]
    
    \includegraphics[width=57pt,height=57pt]{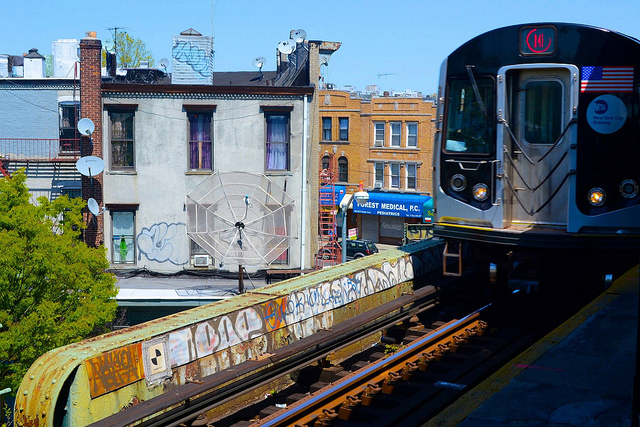} & \includegraphics[width=57pt,height=57pt]{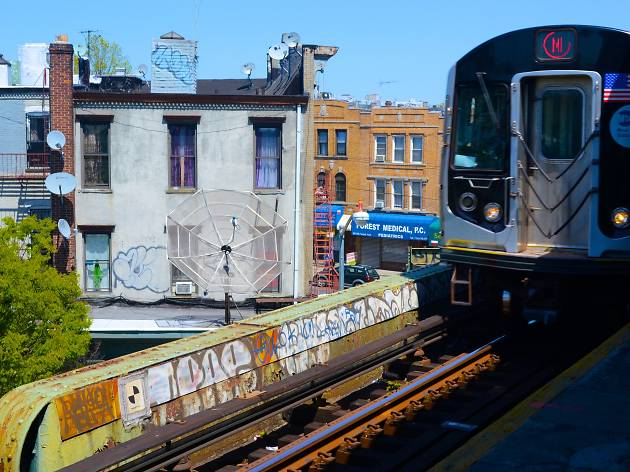} & \includegraphics[width=57pt,height=57pt]{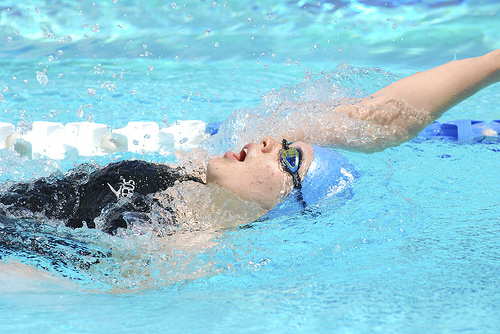} & \includegraphics[width=57pt,height=57pt]{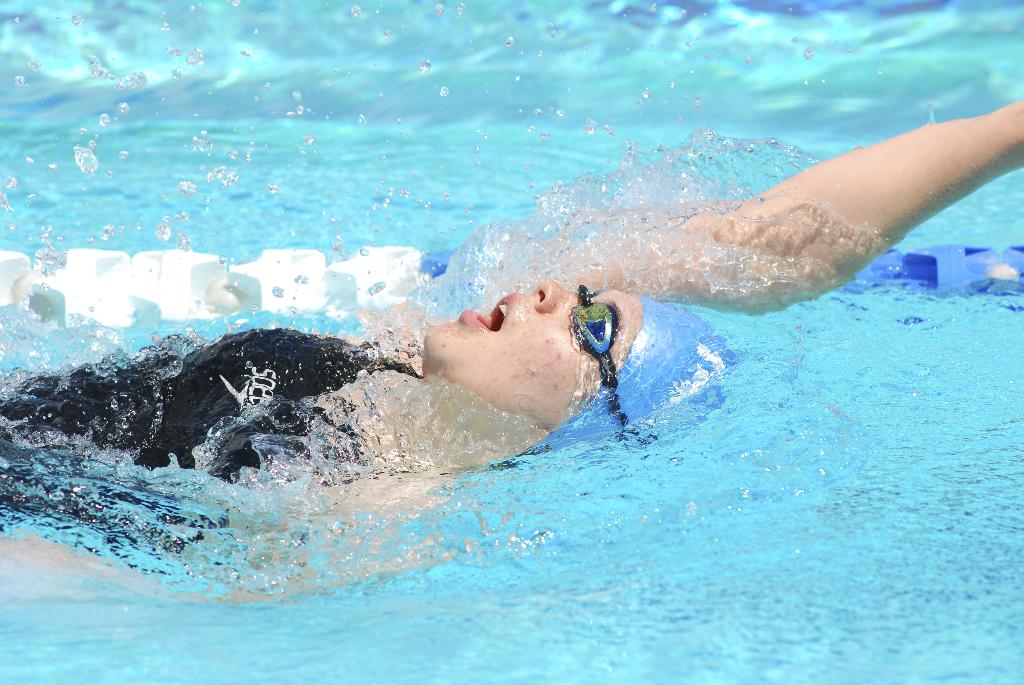}\\ [10pt]
    \end{tabular}
    \captionsetup{justification=centering}
    \subcaption{hard leakage}
    \label{fig:hard_leakage}
\end{subfigure}
\hspace{30pt}
\begin{subfigure}[b]{0.45\linewidth}
    \centering
    \begin{tabular}{@{\hspace{3pt}}c@{\hspace{3pt}}c@{\hspace{8pt}}c@{\hspace{3pt}}c@{\hspace{3pt}}}

    \includegraphics[width=57pt,height=57pt]{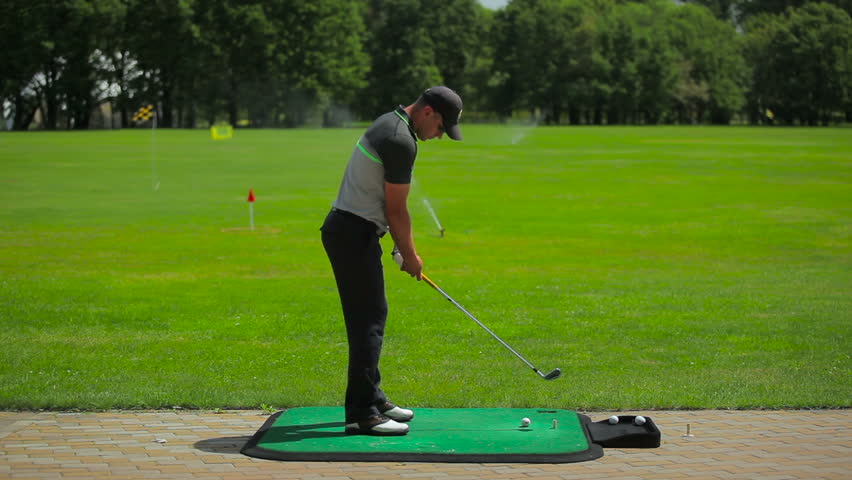} & \includegraphics[width=57pt,height=57pt]{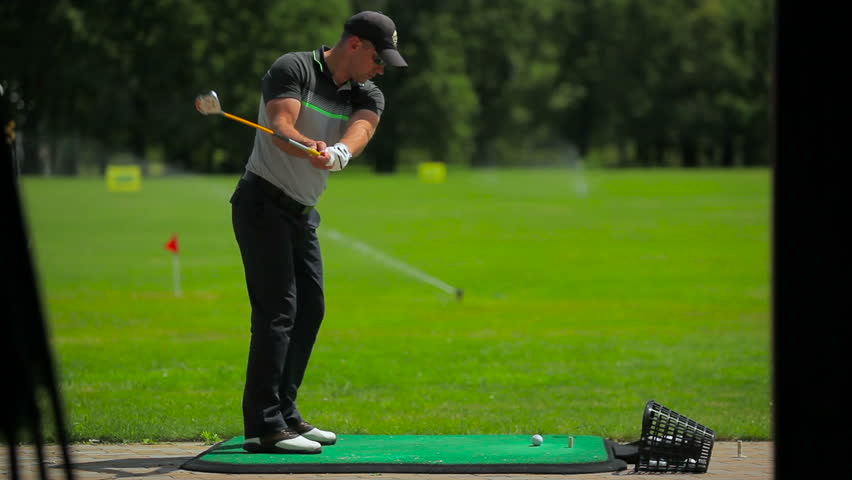} & \includegraphics[width=57pt,height=57pt]{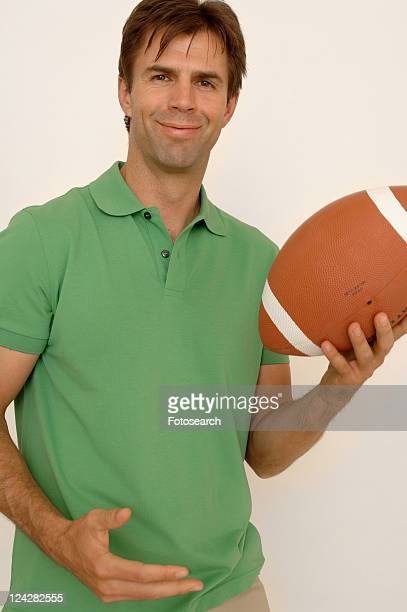} & \includegraphics[width=57pt,height=57pt]{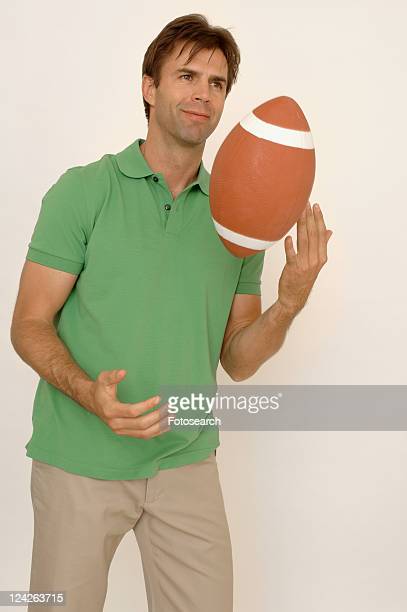}\\[10pt]

    \includegraphics[width=57pt,height=57pt]{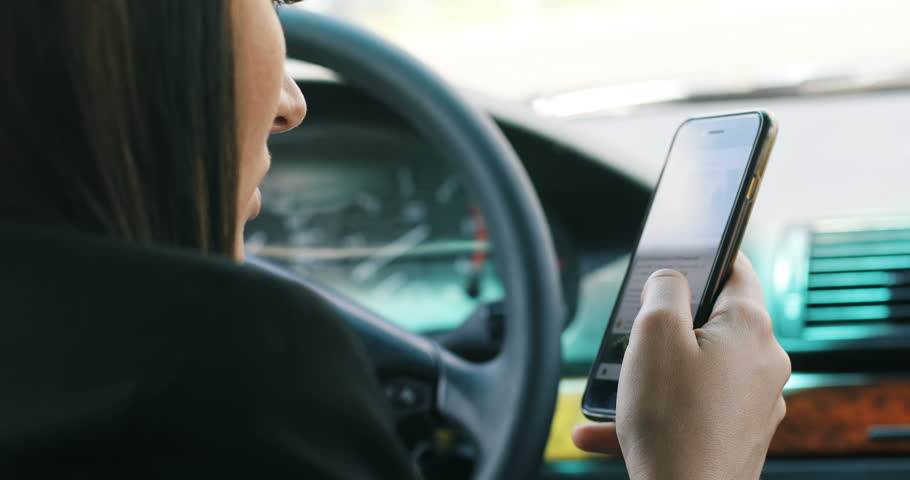} & \includegraphics[width=57pt,height=57pt]{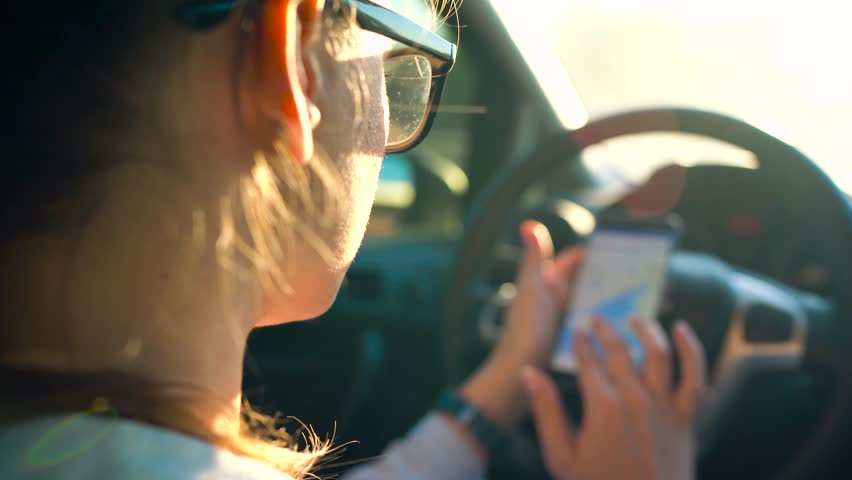} & 
     \includegraphics[width=57pt,height=57pt]{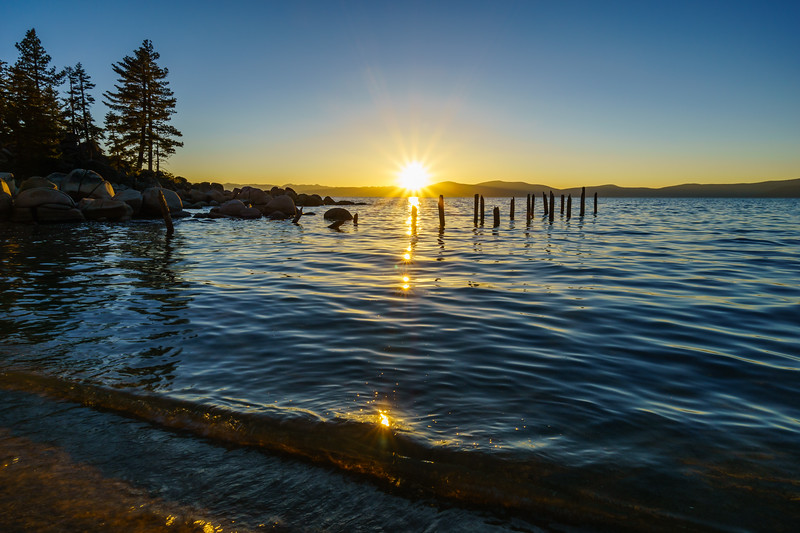} & \includegraphics[width=57pt,height=57pt]{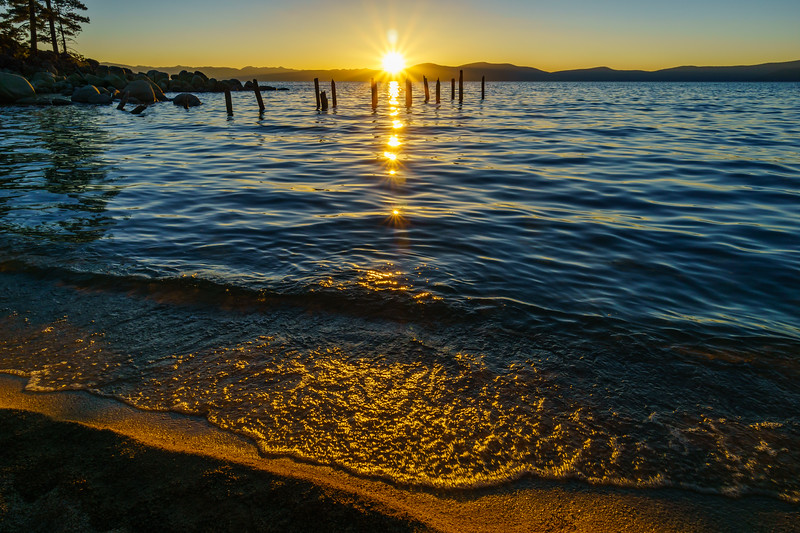}\\ [10pt]
    \end{tabular}
    \captionsetup{justification=centering}
    \subcaption{soft leakage}
    \label{fig:soft_leakage}
\end{subfigure}
\caption{Examples of leaked images. Left: hard leakage, where images are identical, both within the same dataset (first row) and between datasets (second row). Right: soft leakage, where images are near-identical.}
\label{fig:examples_main}
\end{figure*}

%% file: tables/openclip.tex
\begin{table}[t]
\centering
\renewcommand{\arraystretch}{1.1}
\setlength{\tabcolsep}{7pt}
\caption{Zero-shot classification accuracy on ImageNet val split for different subsets, with and without leakage. The last column (gain) indicates the difference with respect to the original dataset (\ie, first row). We highlight the rows of the leaked subsets.}

\vspace{-3pt}
\resizebox{0.82\columnwidth}{!}{
\begin{tabular}{l r r r r}
\toprule
subset & leakage & images & openclip & gain \\
\midrule 

original & - & $50,000$ & $54.18$ & - \\ 
\midrule 

non-leaked  & hard & $49,729$ & $54.15$ & \textcolor{red}{\scriptsize $-0.03$} \\ 
\rowcolor{lighttomato}
leaked  & hard & $271$ & \textbf{$\textbf{60.15}$} & \textcolor{teal}{\scriptsize $+5.97$}\\
random  & -& $271$ & $53.51$ & \textcolor{red}{\scriptsize $-0.67$}\\ 
\midrule 

non-leaked  & soft & $2,281$ & $53.54$ & \textcolor{red}{\scriptsize $-0.64$}\\ 
\rowcolor{lightsteelblue}
leaked  & soft & $2,281$ & \textbf{$\textbf{67.65}$} & \textcolor{teal}{\scriptsize $+13.47$} \\
random  & - & $2,281$ & $54.84$ & \textcolor{teal}{\scriptsize $+0.66 $}\\

\bottomrule
\end{tabular}
}
\label{tab:openclip}
\end{table}

%% file: tables/resnet.tex
\begin{table}[t]
\centering
\renewcommand{\arraystretch}{1.1}
\setlength{\tabcolsep}{3pt}
\caption{Supervised image classification accuracy on ImageNet val split for different subsets, with and without leakage. \textit{Gain} columns indicate the difference with respect to the original dataset (\ie, first row). We highlight the rows of the leaked subsets.}

\vspace{-3pt}
\resizebox{\columnwidth}{!}{
\begin{tabular}{@{}l r r r r r r}
\toprule
subset & leakage & images & resnet50 & gain & resnet152 & gain  \\
\midrule 

original & - & $50,000$ & $80.37$ & - & $82.83$ & - \\ 
\midrule 

non-leaked  & hard & $49,211$ & $81.18$ & \textcolor{teal}{\scriptsize $+0.81$} & $83.65$ & \textcolor{teal}{\scriptsize $+0.82$} \\ 
\rowcolor{lighttomato}
leaked  & hard & $789$ & $30.16$ & \textcolor{red}{\scriptsize $-50.21$} & $31.69$ & \textcolor{red}{\scriptsize $-51.14$} \\
\rowcolor{lighttomato}
$\hookrightarrow$ same label  & hard & $11$ & $\textbf{100.00}$ & \textcolor{teal}{\scriptsize $+19.63$} & $\textbf{100.00}$ & \textcolor{teal}{\scriptsize $+17.17$} \\
\rowcolor{lighttomato}
$\hookrightarrow$ different label  & hard & $778$ & $29.18$ & \textcolor{red}{\scriptsize $-51.19$} & $30.72$ & \textcolor{red}{\scriptsize $-52.11$} \\
random  & -& $789$ & $83.27$ & \textcolor{teal}{\scriptsize $+2.90$} & $84.54$ & \textcolor{teal}{\scriptsize $+1.71$} \\ 
\midrule 

non-leaked  & soft & $36,720$ & $81.00$ & \textcolor{teal}{\scriptsize $+0.63$} & $83.52$ & \textcolor{teal}{\scriptsize $+0.69$} \\
\rowcolor{lightsteelblue}
leaked  & soft & $1,680$ & $62.44$ & \textcolor{red}{\scriptsize $-17.93$} & $62.80$ & \textcolor{red}{\scriptsize $-20.03$} \\
\rowcolor{lightsteelblue}
$\hookrightarrow$ same label  & soft & $812$ & $\textbf{96.06}$ & \textcolor{teal}{\scriptsize $+15.06$} & $\textbf{95.44}$ & \textcolor{teal}{\scriptsize $+11.92$}  \\
\rowcolor{lightsteelblue}
$\hookrightarrow$ different label  & soft & $868$ & $30.99$ & \textcolor{red}{\scriptsize $-50.01$} & $32.26$ & \textcolor{red}{\scriptsize $-51.26$} \\
random  & - & $1,680$ & $80.24$ & \textcolor{red}{\scriptsize $-0.13$} & $83.04$ & \textcolor{teal}{\scriptsize $+0.21$} \\

\bottomrule
\end{tabular}
}
\label{tab:resnet}
\end{table}

%% file: tables/retrieval.tex
\begin{table}[t]
\centering
\renewcommand{\arraystretch}{1.1}
\setlength{\tabcolsep}{5pt}
\caption{Image-to-text retrieval on Flickr30k for different subsets, with and without leakage. Leaked subsets are highlighted.}

\vspace{-3pt}
\resizebox{\columnwidth}{!}{
\begin{tabular}{@{}l r c c c}
\toprule
subset & leakage   & R@1 & R@5 & R@10 \\
\midrule 

original & - & $33.22$ & $55.25$ & $64.13$ \\ 
\midrule 

non-leaked  & hard &  $36.00 \pm 3.14$ & $58.65 \pm 4.19$ & $66.15          
             \pm 2.96$\\ 
\rowcolor{lighttomato}
leaked  & hard &  $\textbf{45.55} \pm 1.19$ & $\textbf{70.80} \pm 1.11$ & $\textbf{79.20} \pm 0.67$\\
random  & - &  $34.30 \pm 2.85 $ & $55.95 \pm 2.54$ & $64.80 \pm 2.67$\\ 
\midrule 

non-leaked  & soft &  $33.05 \pm 3.56$ & $54.95 \pm 3.12$ & $63.75 \pm     2.15$\\
\rowcolor{lightsteelblue}
leaked  & soft & $\textbf{34.90} \pm 2.07$ & $\textbf{59.05} \pm 2.76$ & $\textbf{68.25} \pm 2.12$\\
random  & -  &  $32.15 \pm 3.56$ & $56.35 \pm 5.18$ & $64.25 \pm 4.21$ \\

\bottomrule
\end{tabular}
}
\label{tab:retrieval}
\end{table}

%% file: sec/8_conclusion.tex
\section{Final remarks}
Data leakage is a widespread issue, prevalent in most visual datasets. Leakage can obscure the generalization ability of models, which is particularly problematic when comparing models trained on different datasets, leading to unfair comparisons. We urge dataset designers to carefully consider the implications of these evaluations. For a fairer model evaluation, we recommend the use of duplicate detectors that considers both hard and soft leakage. Ideally, leaked images should be removed from the training set, and if not possible, they should at least be removed from the test set.

\section{Conclusion}
We thoroughly investigated data leakage in visual datasets. We started by characterizing visual leakage, and proposed a data leakage detector based on CLIP image retrieval. We demonstrated the presence of data leakage not only between different datasets (inter-dataset) but also within splits of the same dataset (intra-dataset). Furthermore, we analyzed the impact of data leakage on three downstream tasks, and showed that models not only do memorize identical samples (hard leakage), but also near-identical ones (soft leakage). Overall, we showed consistent evidence that leakage poses a serious threat to fair model evaluation in visual datasets, compromising one of the most fundamental machine learning principles: to not evaluate models on their training data.

%% file: sec/X_suppl.tex
\clearpage

\setcounter{page}{1}
\maketitlesupplementary

\input{figures/transformations}

\input{figures/leakage-examples-supp}

%% file: figures/transformations.tex
\noindent\begin{minipage}{\textwidth}
  \centering
\begin{minipage}[t]{0.17\linewidth}
    \centering
    \includegraphics[width=2.9cm]{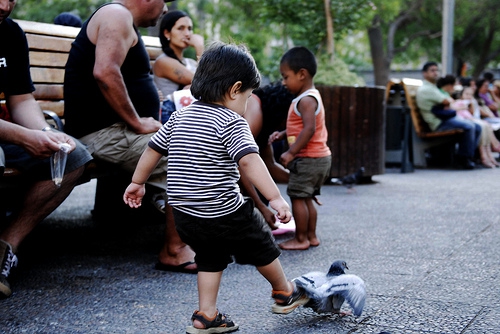}
    \captionsetup{justification=centering}
    {\small original}
\end{minipage}
\begin{minipage}[t]{0.17\linewidth}
    \centering
    \includegraphics[width=2.9cm]{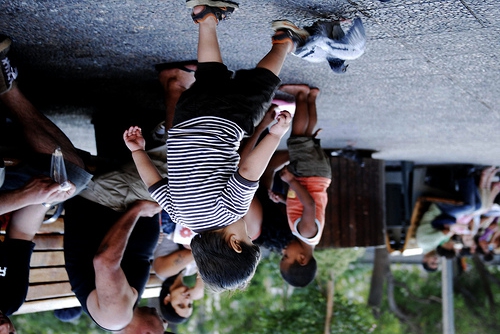}
    \captionsetup{justification=centering}
    {\small flip-v}
\end{minipage}
\begin{minipage}[t]{0.17\linewidth}
    \centering
    \includegraphics[width=2.9cm]{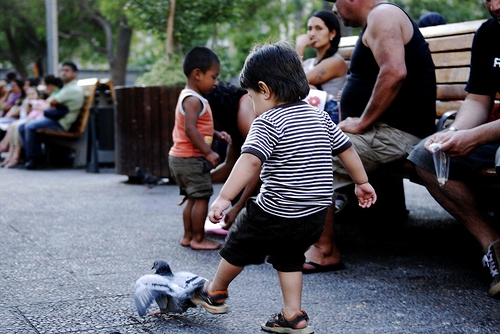}
    \captionsetup{justification=centering}
    {\small flip-h}
\end{minipage}
\begin{minipage}[t]{0.17\linewidth}
    \centering
    \includegraphics[width=2.9cm]{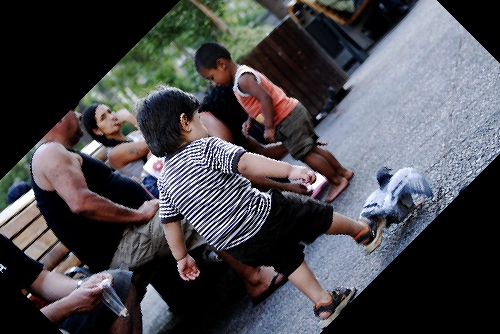}
    \captionsetup{justification=centering}
    {\small rotation-45}
\end{minipage}
\begin{minipage}[t]{0.17\linewidth}
    \centering
    \includegraphics[width=2.9cm]{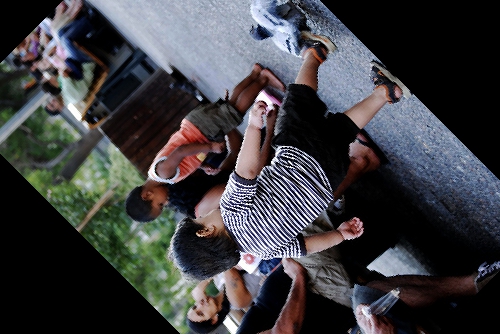}
    \captionsetup{justification=centering}
    {\small rotation-135}
\end{minipage}
\vspace{7pt}

\begin{minipage}[t]{0.17\linewidth}
    \centering
    \includegraphics[width=2.9cm]{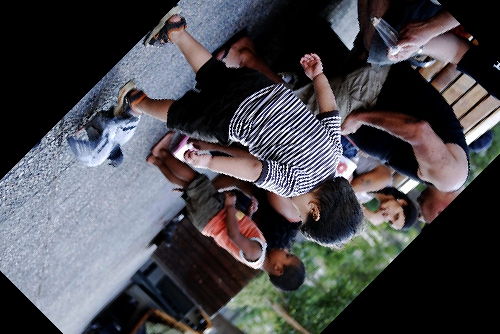}
    \captionsetup{justification=centering}
    {\small rotation-225}
\end{minipage}
\begin{minipage}[t]{0.17\linewidth}
    \centering
    \includegraphics[width=2.9cm]{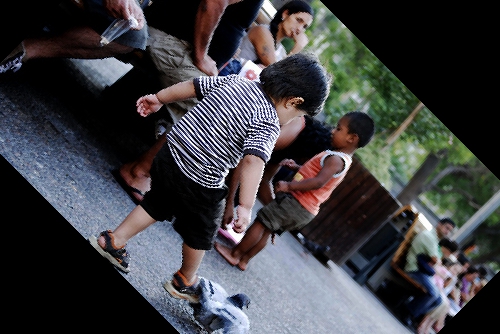}
    \captionsetup{justification=centering}
    {\small rotation-315}
\end{minipage}
\begin{minipage}[t]{0.17\linewidth}
    \centering
    \includegraphics[width=2.9cm]{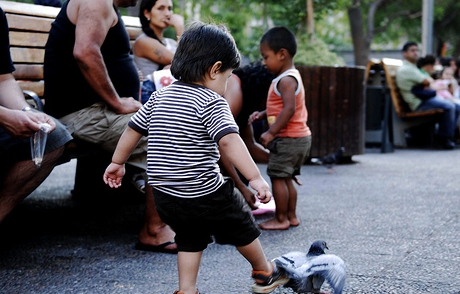}
    \captionsetup{justification=centering}
    {\small crop-20}
\end{minipage}
\begin{minipage}[t]{0.17\linewidth}
    \centering
    \includegraphics[width=2.9cm]{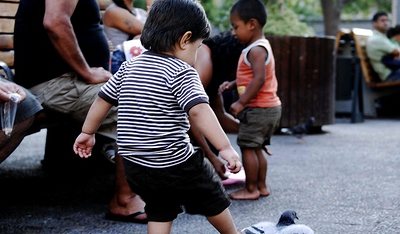}
    \captionsetup{justification=centering}
    {\small crop-50}
\end{minipage}
\begin{minipage}[t]{0.17\linewidth}
    \centering
    \includegraphics[width=2.9cm]{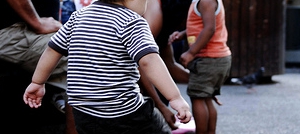}
    \captionsetup{justification=centering}
    {\small crop-100}
\end{minipage}
\vspace{7pt}

\begin{minipage}[t]{0.17\linewidth}
    \centering
    \includegraphics[width=2.9cm]{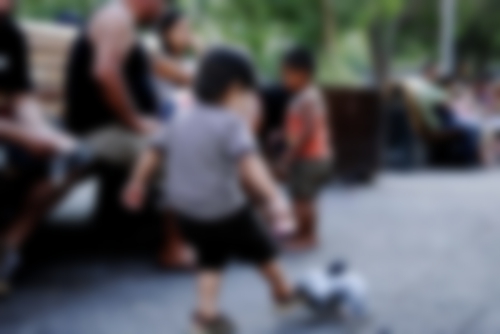}
    \captionsetup{justification=centering}
    {\small gaussian-filter}
\end{minipage}
\begin{minipage}[t]{0.17\linewidth}
    \centering
    \includegraphics[width=2.9cm]{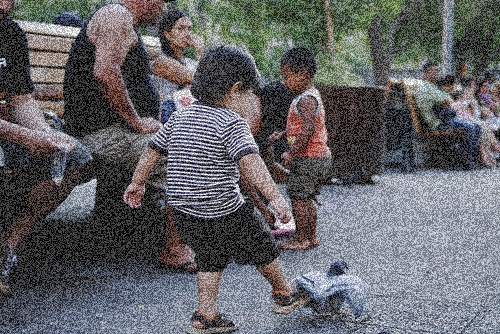}
    \captionsetup{justification=centering}
    {\small noise}
\end{minipage}
\begin{minipage}[t]{0.17\linewidth}
    \centering
    \includegraphics[width=2.9cm]{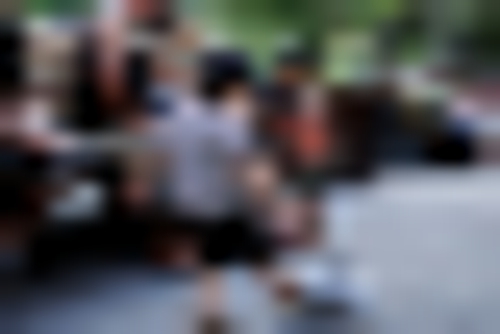}
    \captionsetup{justification=centering}
    {\small resize-20}
\end{minipage}
\begin{minipage}[t]{0.17\linewidth}
    \centering
    \includegraphics[width=2.9cm]{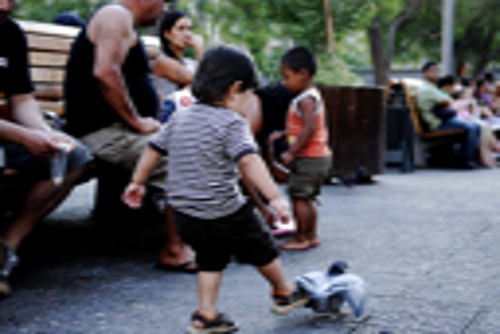}
    \captionsetup{justification=centering}
    {\small resize-128}
\end{minipage}
\begin{minipage}[t]{0.17\linewidth}
    \centering
    \includegraphics[width=2.9cm]{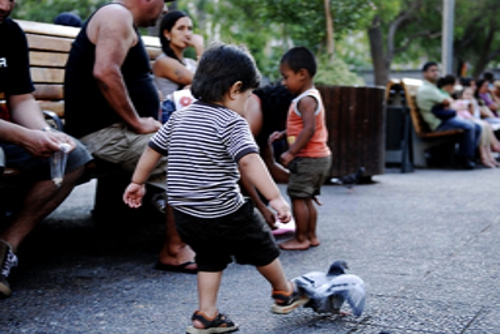}
    \captionsetup{justification=centering}
    {\small resize-256}
\end{minipage}
\vspace{7pt}

\begin{minipage}[t]{0.17\linewidth}
    \centering
    \includegraphics[width=2.9cm]{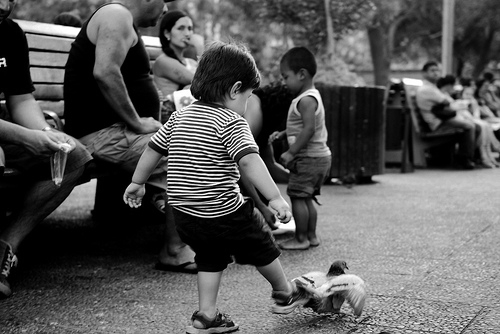}
    \captionsetup{justification=centering}
    {\small grayscale}
\end{minipage}
\begin{minipage}[t]{0.17\linewidth}
    \centering
    \includegraphics[width=2.9cm]{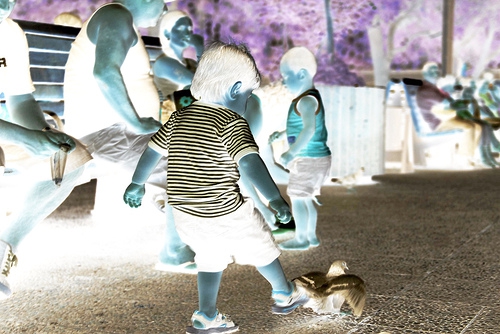}
    \captionsetup{justification=centering}
    {\small invert}
\end{minipage}
\begin{minipage}[t]{0.17\linewidth}
    \centering
    \includegraphics[width=2.9cm]{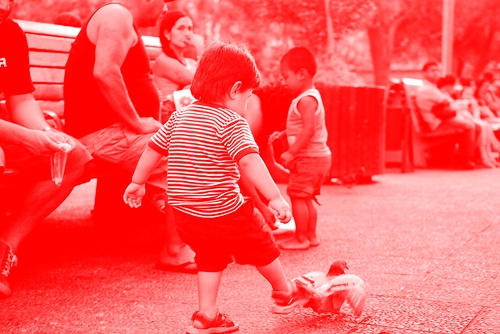}
    \captionsetup{justification=centering}
    {\small red}
\end{minipage}
\begin{minipage}[t]{0.17\linewidth}
    \centering
    \includegraphics[width=2.9cm]{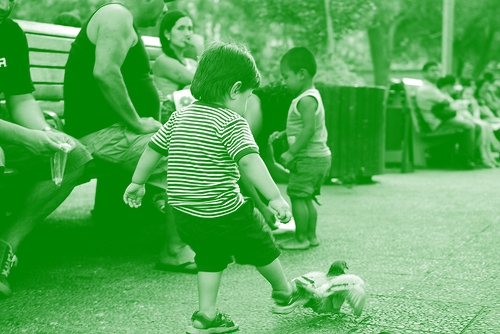}
    \captionsetup{justification=centering}
    {\small green}
\end{minipage}
\begin{minipage}[t]{0.17\linewidth}
    \centering
    \includegraphics[width=2.9cm]{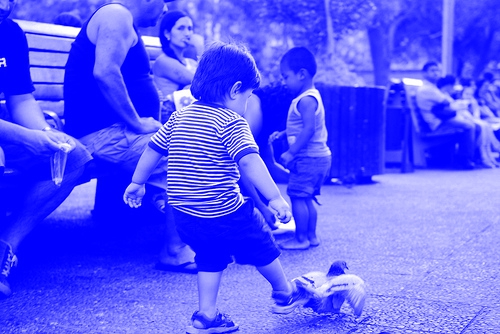}
    \captionsetup{justification=centering}
    {\small blue}
\end{minipage} %
\vspace{7pt}

\captionof{figure}{Transformations used in Section 6 for method evaluation.}
\label{fig:transformation}
\end{minipage}

%% file: figures/leakage-examples-supp.tex
\begin{figure*}[tb]
  \centering
\hspace{-25pt}
\centering
\begin{tabular}{@{\hspace{3pt}}c@{\hspace{3pt}}c@{\hspace{8pt}}c@{\hspace{3pt}}c@{\hspace{8pt}}c@{\hspace{3pt}}c@{\hspace{8pt}}c@{\hspace{3pt}}c@{\hspace{3pt}}}

\includegraphics[width=57pt,height=57pt]{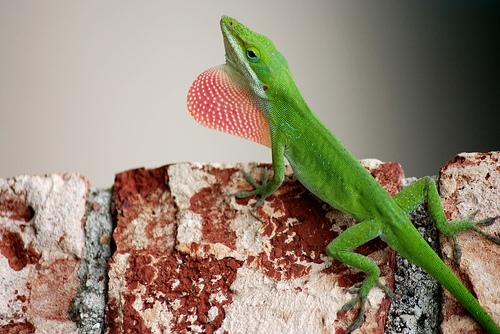} & \includegraphics[width=57pt,height=57pt]{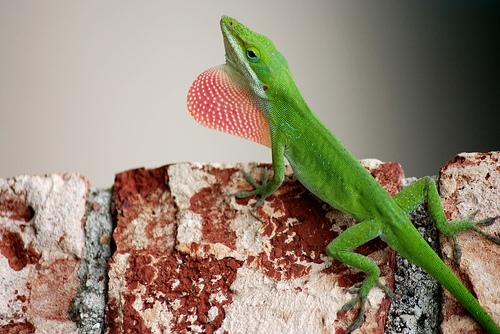} & \includegraphics[width=57pt,height=57pt]{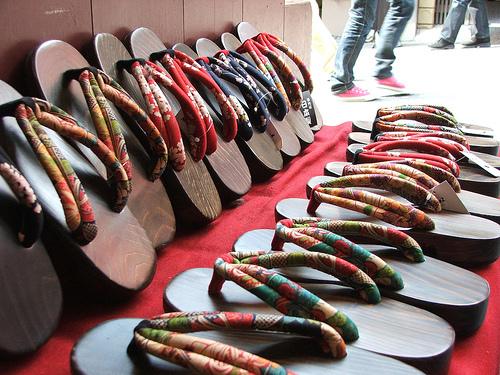} & \includegraphics[width=57pt,height=57pt]{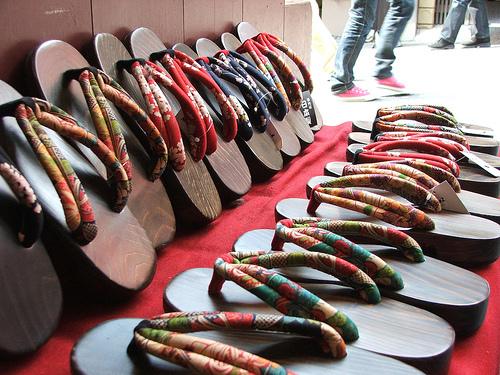} & \includegraphics[width=57pt,height=57pt]{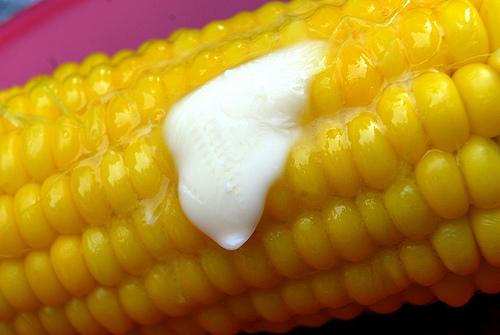} & \includegraphics[width=57pt,height=57pt]{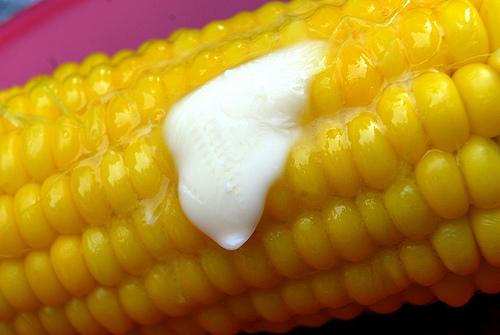} & \includegraphics[width=57pt,height=57pt]{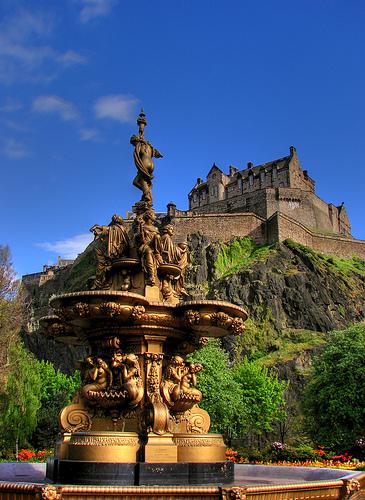} & \includegraphics[width=57pt,height=57pt]{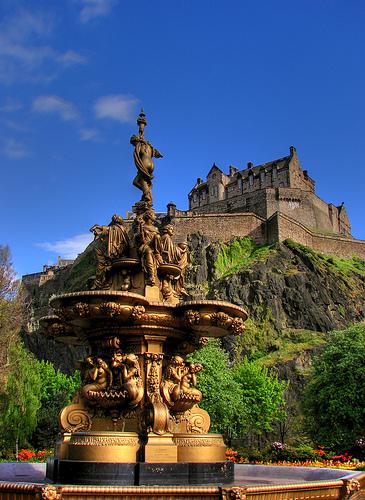} \\ 
\multicolumn{2}{c}{\scriptsize 100.00} & \multicolumn{2}{c}{\scriptsize 100.00} & \multicolumn{2}{c}{\scriptsize 100.00} & \multicolumn{2}{c}{\scriptsize 100.00}\\[5pt]
\includegraphics[width=57pt,height=57pt]{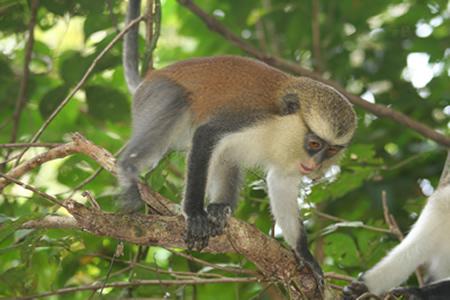} & \includegraphics[width=57pt,height=57pt]{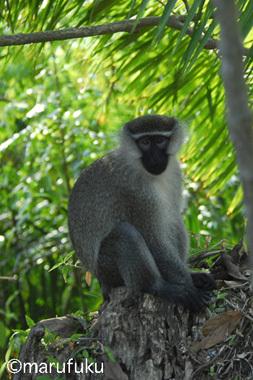} & \includegraphics[width=57pt,height=57pt]{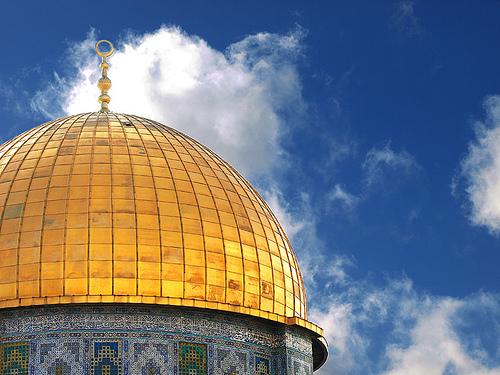} & \includegraphics[width=57pt,height=57pt]{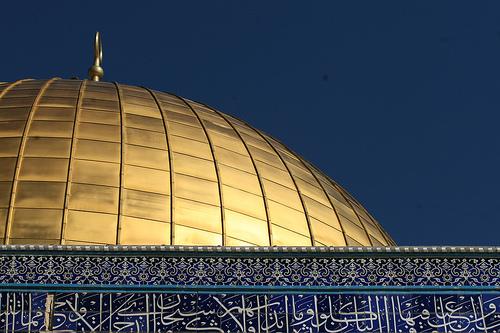} & \includegraphics[width=57pt,height=57pt]{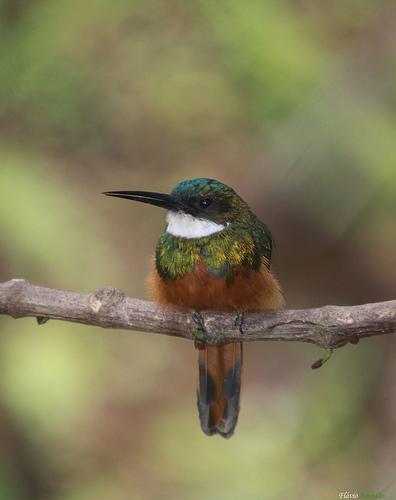} & \includegraphics[width=57pt,height=57pt]{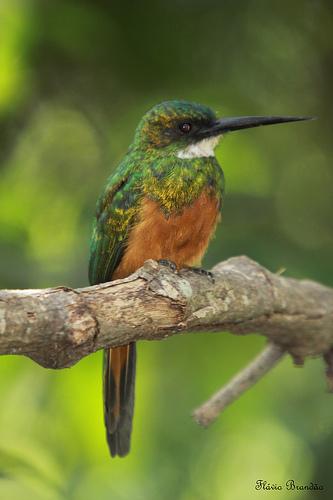} & \includegraphics[width=57pt,height=57pt]{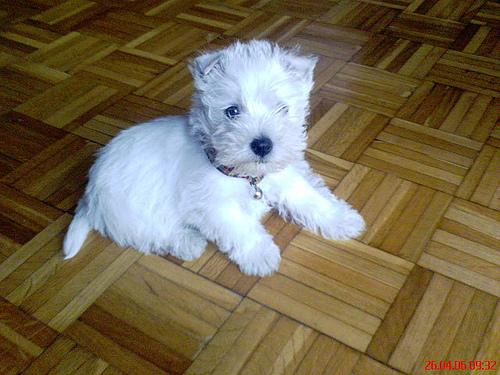} & \includegraphics[width=57pt,height=57pt]{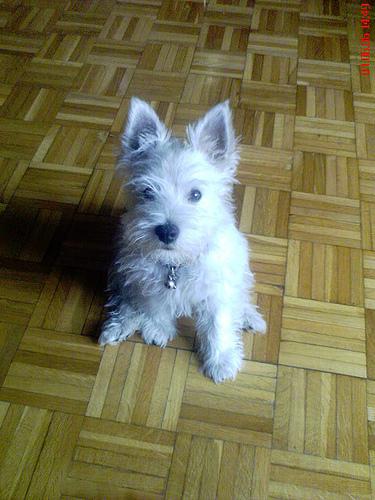} \\ 
\multicolumn{2}{c}{\scriptsize 95.44} & \multicolumn{2}{c}{\scriptsize 95.76} & \multicolumn{2}{c}{\scriptsize 96.73} & \multicolumn{2}{c}{\scriptsize 95.85}\\[5pt]
\includegraphics[width=57pt,height=57pt]{images/supp_examples/coco_val_2014_coco_train_2014_hard_0_query.jpg} & \includegraphics[width=57pt,height=57pt]{images/supp_examples/coco_val_2014_coco_train_2014_hard_0_collection.jpg} & \includegraphics[width=57pt,height=57pt]{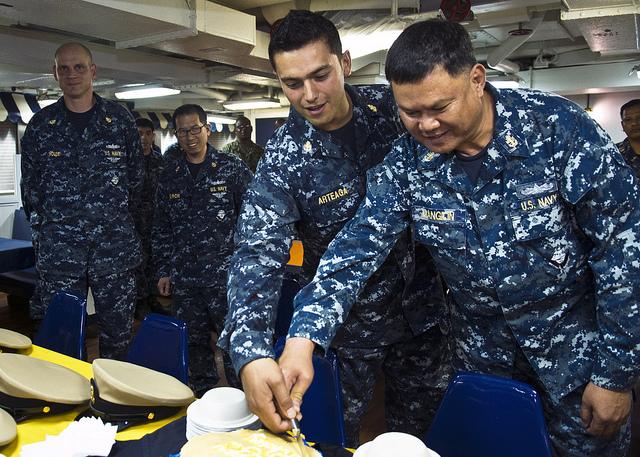} & \includegraphics[width=57pt,height=57pt]{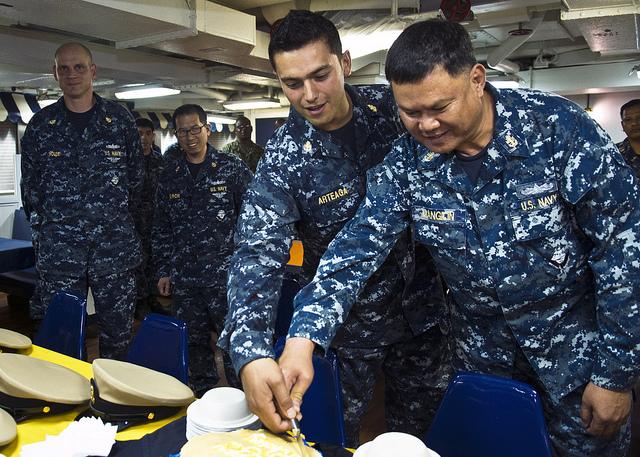} & \includegraphics[width=57pt,height=57pt]{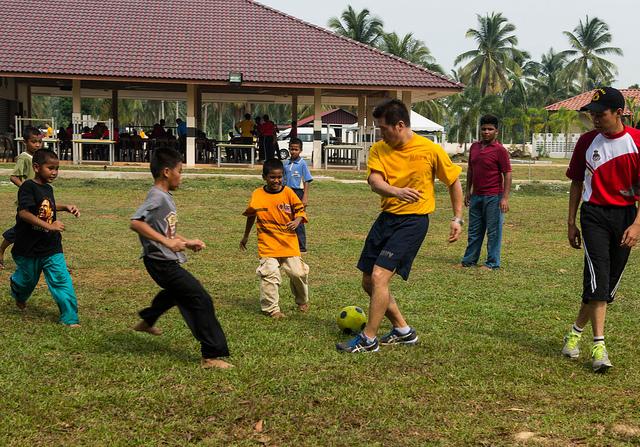} & \includegraphics[width=57pt,height=57pt]{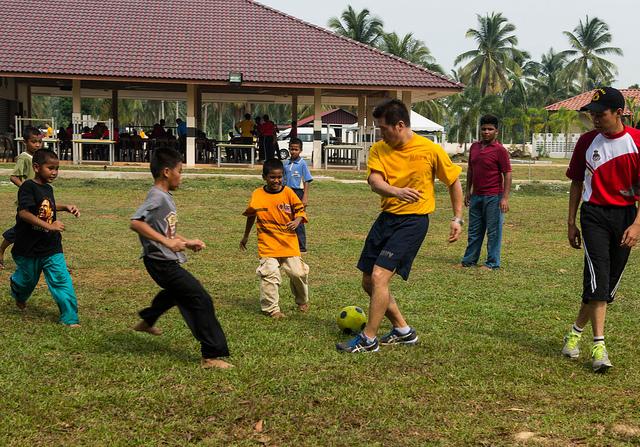} & \includegraphics[width=57pt,height=57pt]{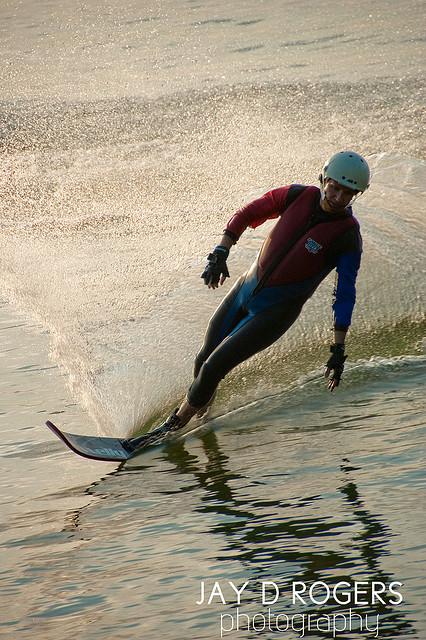} & \includegraphics[width=57pt,height=57pt]{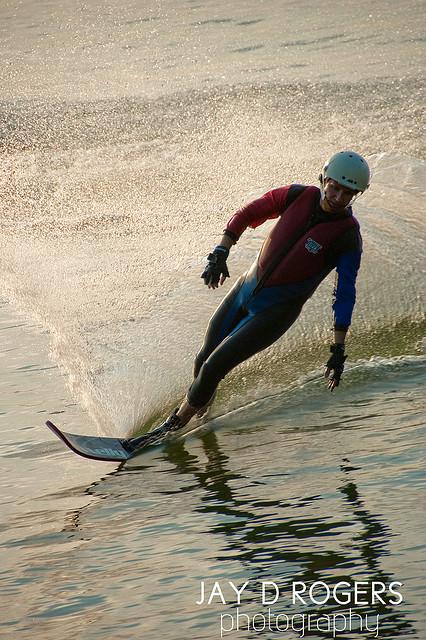} \\ 
\multicolumn{2}{c}{\scriptsize 100.00} & \multicolumn{2}{c}{\scriptsize 100.00} & \multicolumn{2}{c}{\scriptsize 100.00} & \multicolumn{2}{c}{\scriptsize 99.94}\\[5pt]
\includegraphics[width=57pt,height=57pt]{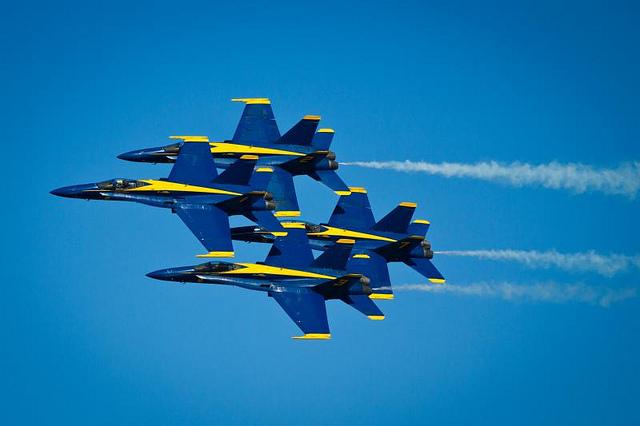} & \includegraphics[width=57pt,height=57pt]{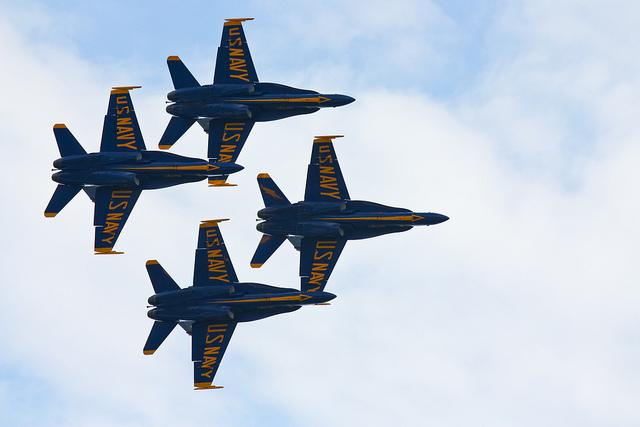} & \includegraphics[width=57pt,height=57pt]{images/supp_examples/coco_val_2014_coco_train_2014_soft_1_query.jpg} & \includegraphics[width=57pt,height=57pt]{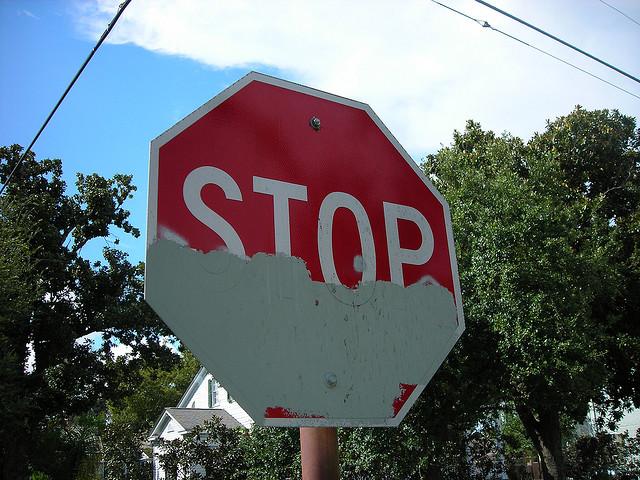} & \includegraphics[width=57pt,height=57pt]{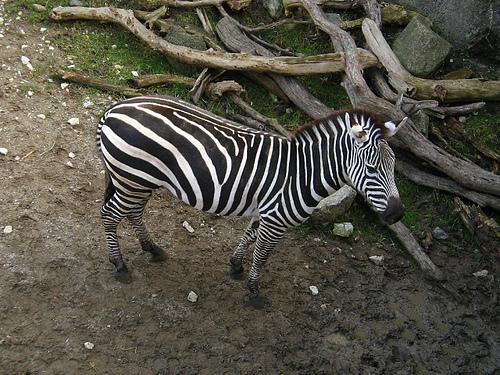} & \includegraphics[width=57pt,height=57pt]{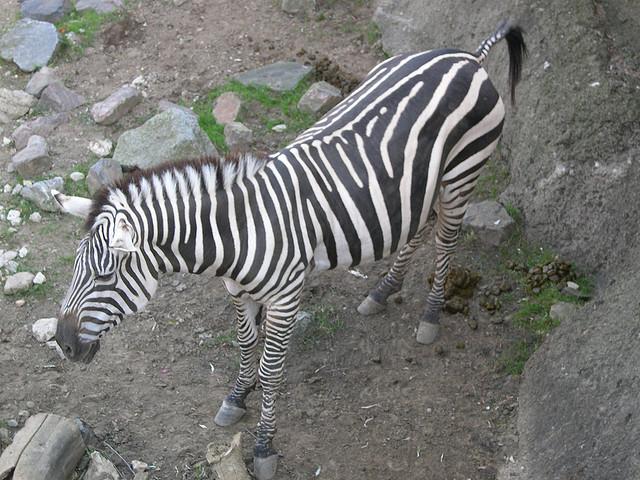} & \includegraphics[width=57pt,height=57pt]{images/supp_examples/coco_val_2014_coco_train_2014_soft_3_query.jpg} & \includegraphics[width=57pt,height=57pt]{images/supp_examples/coco_val_2014_coco_train_2014_soft_3_collection.jpg} \\ 
\multicolumn{2}{c}{\scriptsize 95.02} & \multicolumn{2}{c}{\scriptsize 95.43} & \multicolumn{2}{c}{\scriptsize 95.19} & \multicolumn{2}{c}{\scriptsize 95.05}\\[5pt]
\includegraphics[width=57pt,height=57pt]{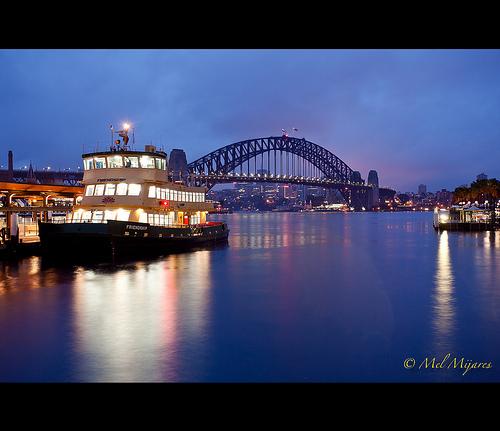} & \includegraphics[width=57pt,height=57pt]{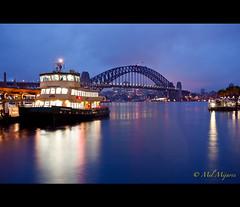} & \includegraphics[width=57pt,height=57pt]{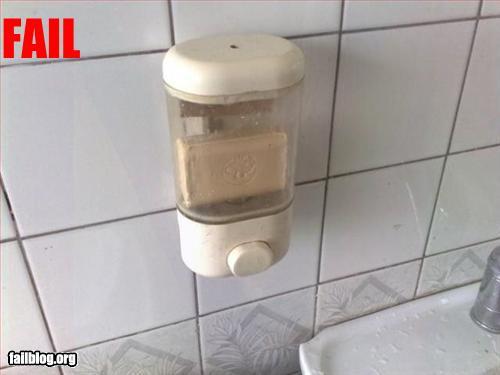} & \includegraphics[width=57pt,height=57pt]{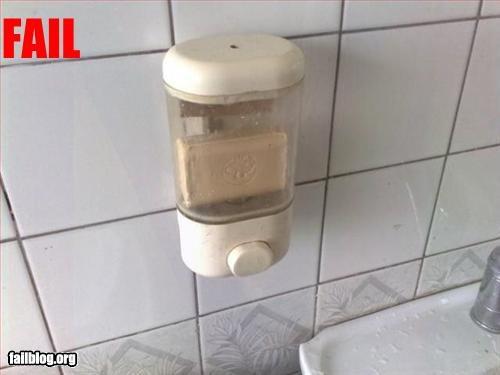} & \includegraphics[width=57pt,height=57pt]{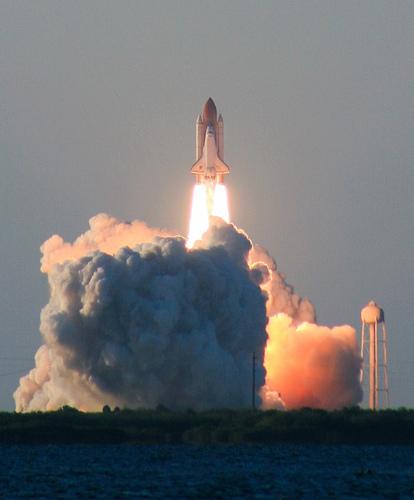} & \includegraphics[width=57pt,height=57pt]{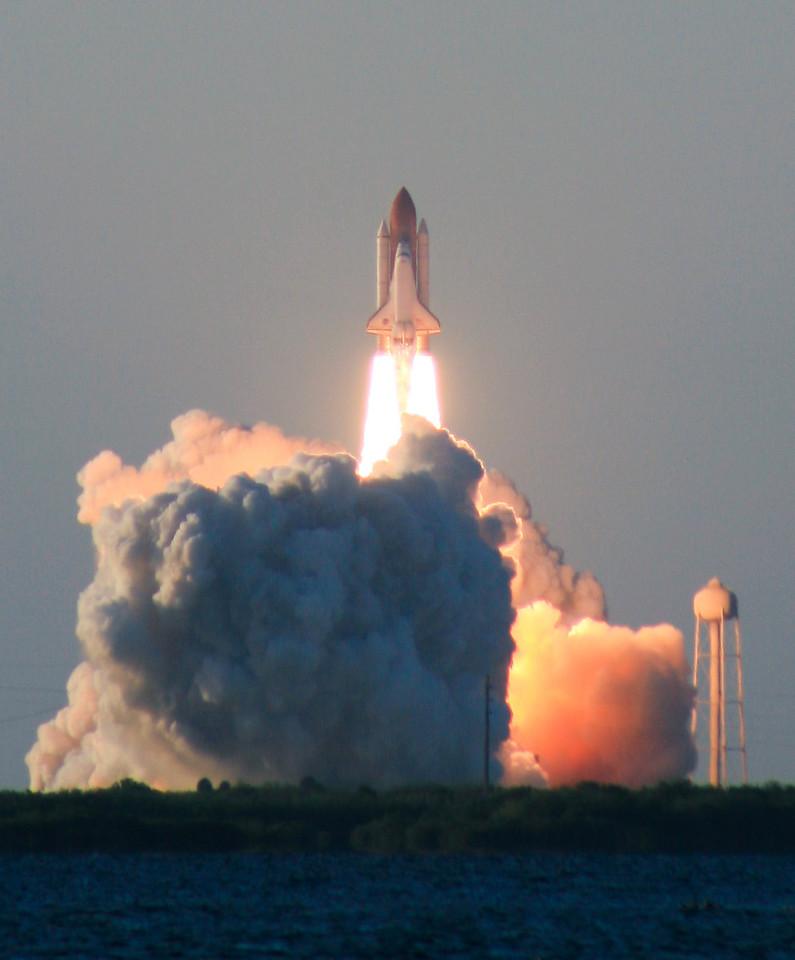} & \includegraphics[width=57pt,height=57pt]{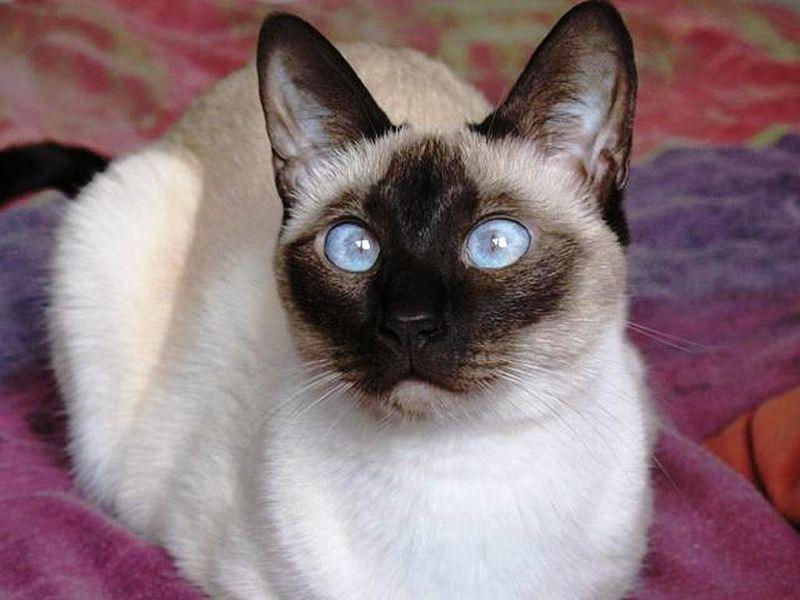} & \includegraphics[width=57pt,height=57pt]{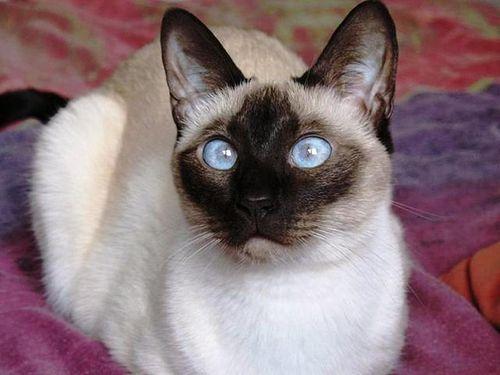} \\ 
\multicolumn{2}{c}{\scriptsize 98.91} & \multicolumn{2}{c}{\scriptsize 98.67} & \multicolumn{2}{c}{\scriptsize 99.10} & \multicolumn{2}{c}{\scriptsize 98.29}\\[5pt]
\includegraphics[width=57pt,height=57pt]{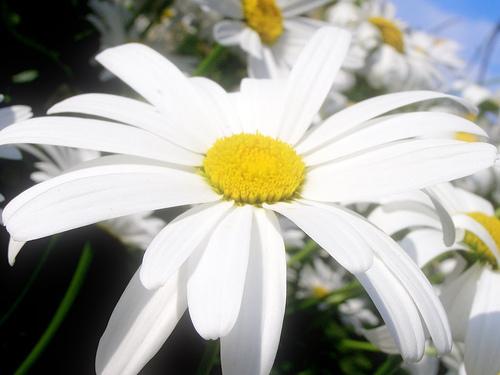} & \includegraphics[width=57pt,height=57pt]{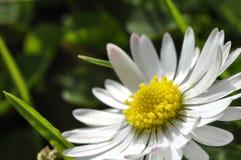} & \includegraphics[width=57pt,height=57pt]{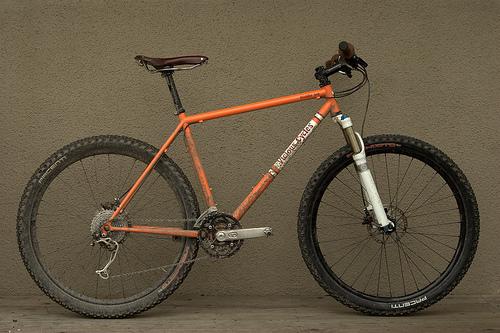} & \includegraphics[width=57pt,height=57pt]{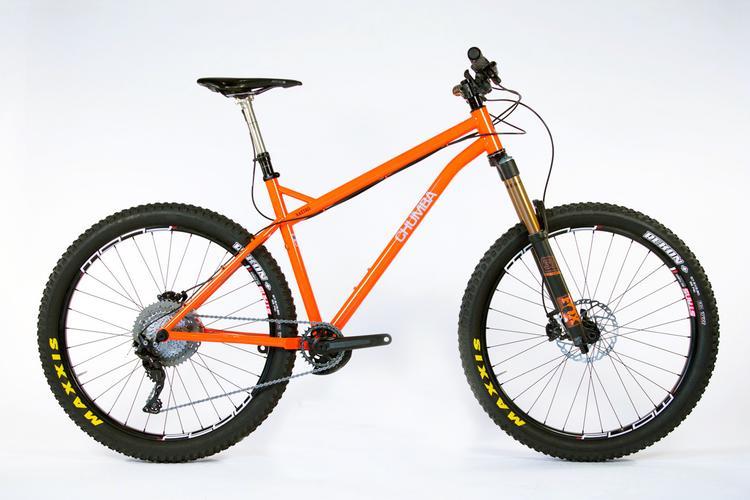} & \includegraphics[width=57pt,height=57pt]{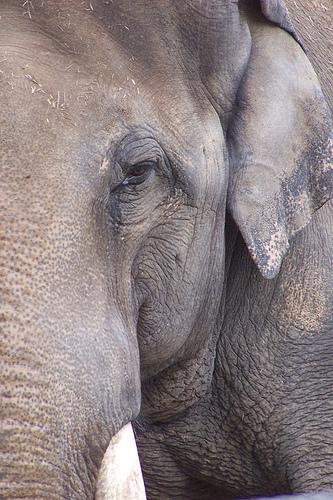} & \includegraphics[width=57pt,height=57pt]{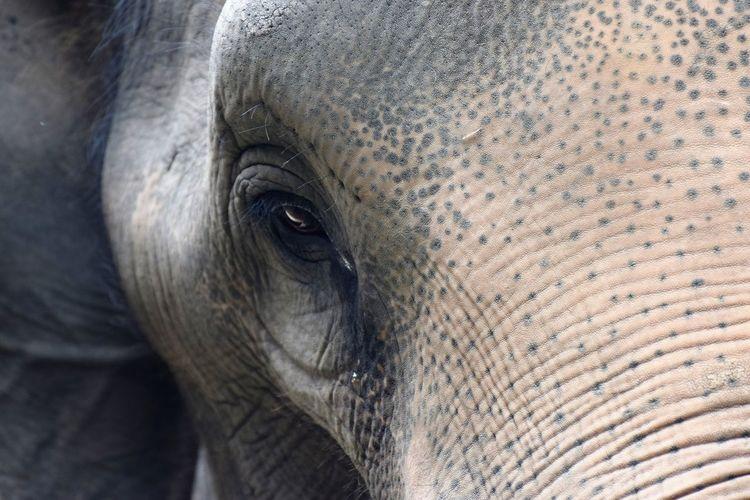} & \includegraphics[width=57pt,height=57pt]{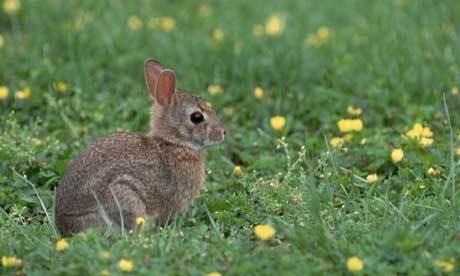} & \includegraphics[width=57pt,height=57pt]{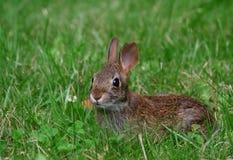} \\ 
\multicolumn{2}{c}{\scriptsize 96.15} & \multicolumn{2}{c}{\scriptsize 96.20} & \multicolumn{2}{c}{\scriptsize 95.54} & \multicolumn{2}{c}{\scriptsize 95.92}\\[5pt]
\includegraphics[width=57pt,height=57pt]{images/supp_examples/coco_val_2014_laion_hard_0_query.jpg} & \includegraphics[width=57pt,height=57pt]{images/supp_examples/coco_val_2014_laion_hard_0_collection.jpg} & \includegraphics[width=57pt,height=57pt]{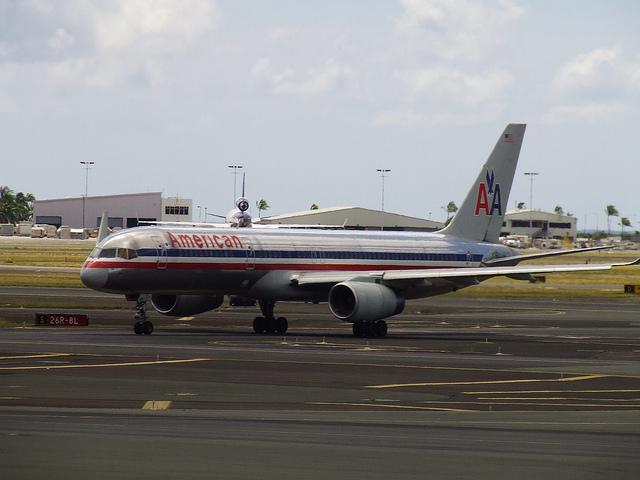} & \includegraphics[width=57pt,height=57pt]{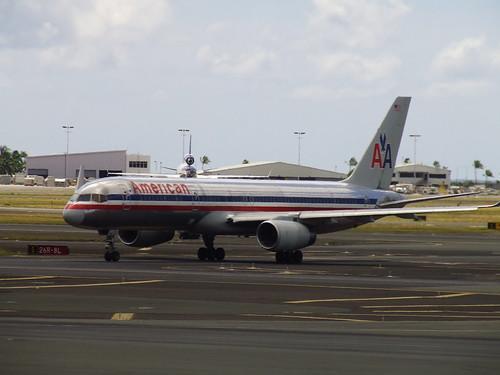} & \includegraphics[width=57pt,height=57pt]{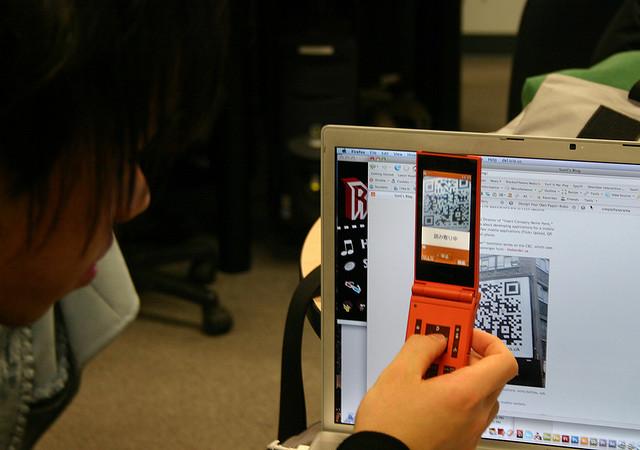} & \includegraphics[width=57pt,height=57pt]{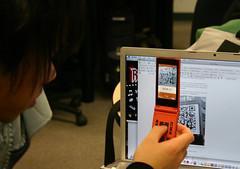} & \includegraphics[width=57pt,height=57pt]{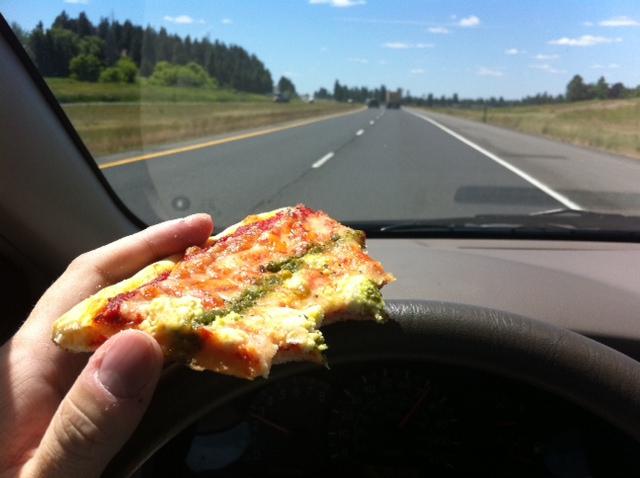} & \includegraphics[width=57pt,height=57pt]{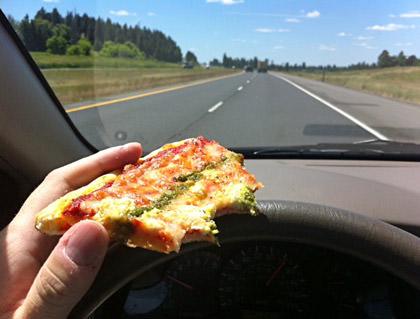} \\ 
\multicolumn{2}{c}{\scriptsize 98.43} & \multicolumn{2}{c}{\scriptsize 98.29} & \multicolumn{2}{c}{\scriptsize 98.23} & \multicolumn{2}{c}{\scriptsize 98.31}\\[5pt]
\includegraphics[width=57pt,height=57pt]{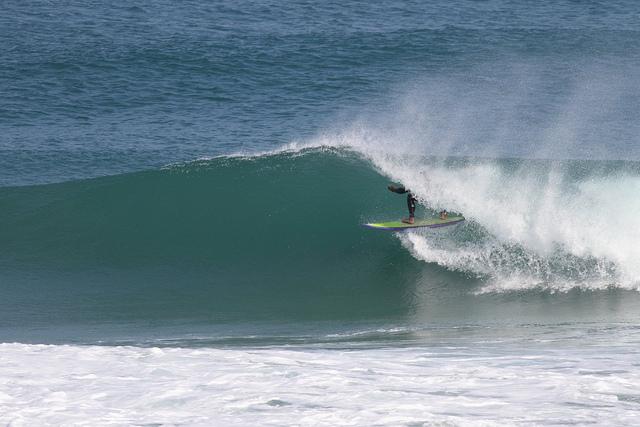} & \includegraphics[width=57pt,height=57pt]{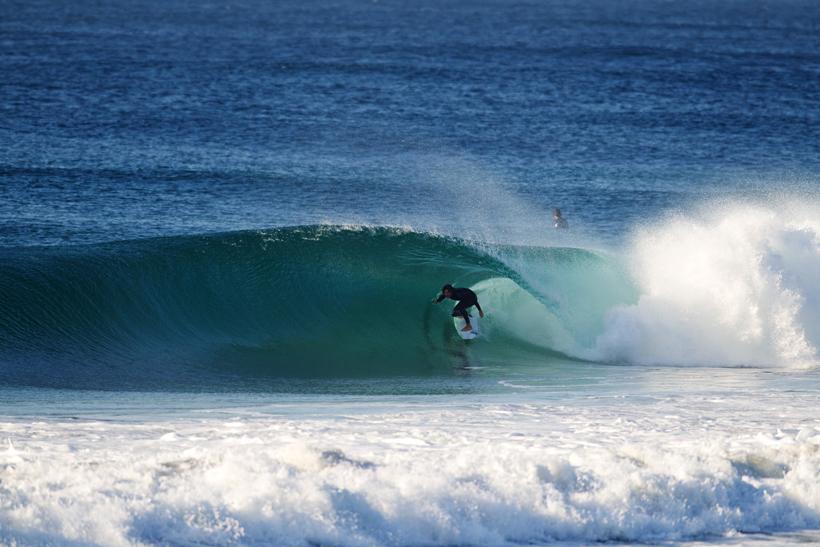} & \includegraphics[width=57pt,height=57pt]{images/supp_examples/coco_val_2014_laion_soft_1_query.jpg} & \includegraphics[width=57pt,height=57pt]{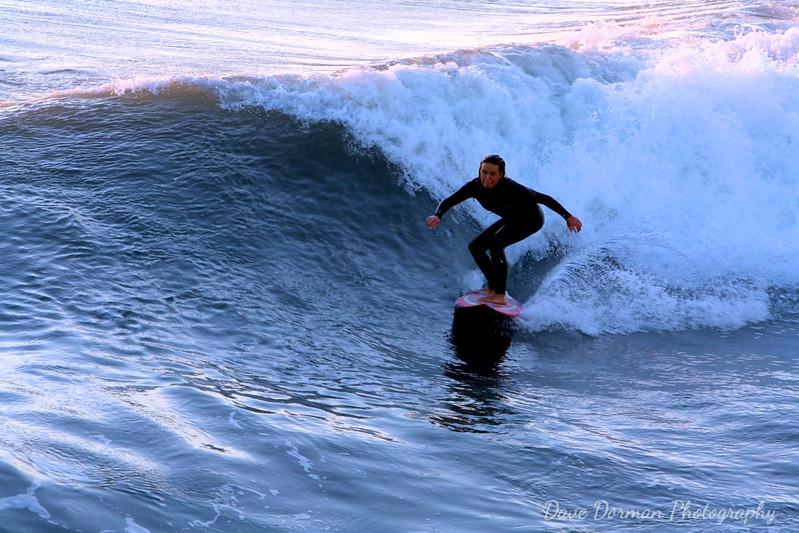} & \includegraphics[width=57pt,height=57pt]{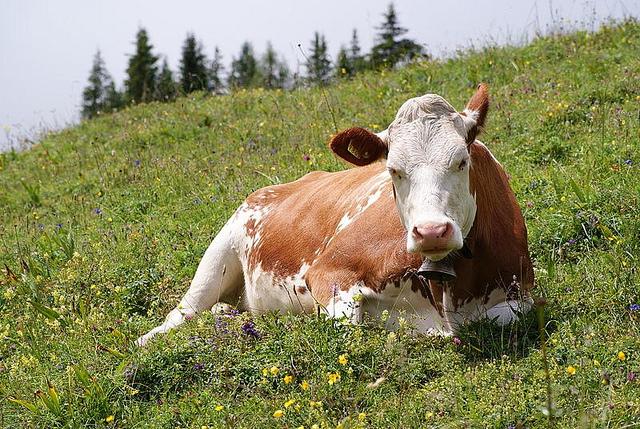} & \includegraphics[width=57pt,height=57pt]{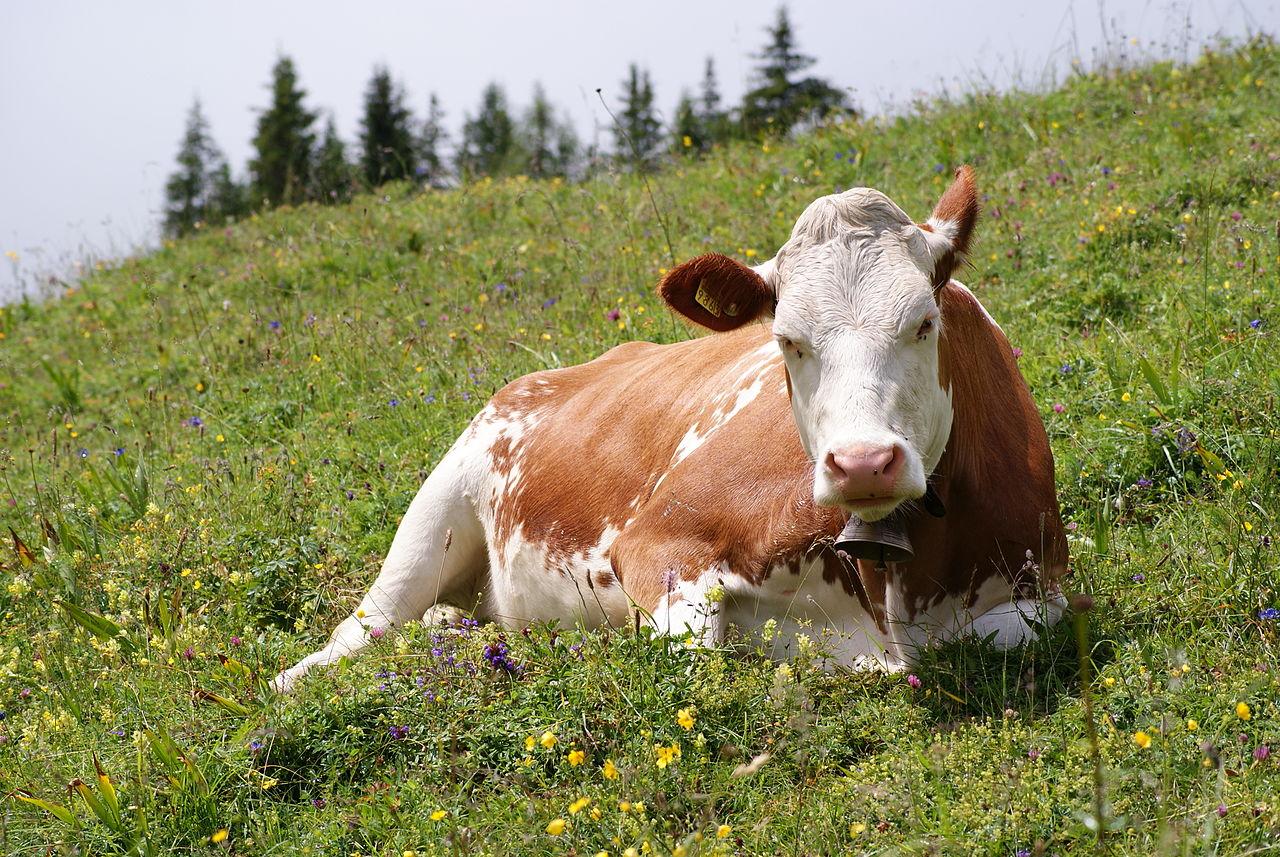} & \includegraphics[width=57pt,height=57pt]{images/supp_examples/coco_val_2014_laion_soft_3_query.jpg} & \includegraphics[width=57pt,height=57pt]{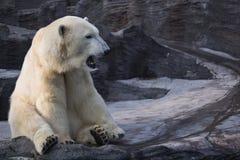} \\ 
\multicolumn{2}{c}{\scriptsize 97.55} & \multicolumn{2}{c}{\scriptsize 96.00} & \multicolumn{2}{c}{\scriptsize 97.23} & \multicolumn{2}{c}{\scriptsize 95.74}\\[5pt]

\end{tabular}
\caption{Examples of leaked images with matching scores.}
\label{fig:examples_main_supp}
\end{figure*}

%% file: main.bbl
\begin{thebibliography}{60}
\providecommand{\natexlab}[1]{#1}
\providecommand{\url}[1]{\texttt{#1}}
\expandafter\ifx\csname urlstyle\endcsname\relax
  \providecommand{\doi}[1]{doi: #1}\else
  \providecommand{\doi}{doi: \begingroup \urlstyle{rm}\Url}\fi

\bibitem[Achiam et~al.(2024)Achiam, Adler, Agarwal, Ahmad, Akkaya, Aleman, Almeida, Altenschmidt, Altman, Anadkat, et~al.]{openai2024gpt4technicalreport}
Josh Achiam, Steven Adler, Sandhini Agarwal, Lama Ahmad, Ilge Akkaya, Florencia~Leoni Aleman, Diogo Almeida, Janko Altenschmidt, Sam Altman, Shyamal Anadkat, et~al.
\newblock {GPT}-4 technical report.
\newblock Technical report, OpenAI, 2024.

\bibitem[Alayrac et~al.(2022)Alayrac, Donahue, Luc, Miech, Barr, Hasson, Lenc, Mensch, Millican, Reynolds, et~al.]{alayrac2022flamingo}
Jean-Baptiste Alayrac, Jeff Donahue, Pauline Luc, Antoine Miech, Iain Barr, Yana Hasson, Karel Lenc, Arthur Mensch, Katherine Millican, Malcolm Reynolds, et~al.
\newblock Flamingo: a visual language model for few-shot learning.
\newblock In \emph{NeurIPS}, 2022.

\bibitem[Anil et~al.(2023)Anil, Dai, Firat, Johnson, Lepikhin, Passos, Shakeri, Taropa, Bailey, Chen, et~al.]{anil2023palm}
Rohan Anil, Andrew~M Dai, Orhan Firat, Melvin Johnson, Dmitry Lepikhin, Alexandre Passos, Siamak Shakeri, Emanuel Taropa, Paige Bailey, Zhifeng Chen, et~al.
\newblock Pa{LM} 2 technical report.
\newblock Technical report, Google, 2023.

\bibitem[Balloccu et~al.(2024)Balloccu, Schmidtová, Lango, and Dušek]{balloccu-etal-2024-leak}
Simone Balloccu, Patrícia Schmidtová, Mateusz Lango, and Ondřej Dušek.
\newblock Leak, cheat, repeat: Data contamination and evaluation malpractices in closed-source {LLMs}.
\newblock In \emph{ACL}, 2024.

\bibitem[Barbu et~al.(2019)Barbu, Mayo, Alverio, Luo, Wang, Gutfreund, Tenenbaum, and Katz]{barbu2019objectnet}
Andrei Barbu, David Mayo, Julian Alverio, William Luo, Christopher Wang, Dan Gutfreund, Josh Tenenbaum, and Boris Katz.
\newblock Object{N}et: A large-scale bias-controlled dataset for pushing the limits of object recognition models.
\newblock In \emph{NeurIPS}, 2019.

\bibitem[Birhane and Prabhu(2021)]{birhane2021large}
Abeba Birhane and Vinay~Uday Prabhu.
\newblock Large image datasets: A pyrrhic win for computer vision?
\newblock In \emph{WACV}, 2021.

\bibitem[Birhane et~al.(2021)Birhane, Prabhu, and Kahembwe]{birhane2021multimodal}
Abeba Birhane, Vinay~Uday Prabhu, and Emmanuel Kahembwe.
\newblock Multimodal datasets: misogyny, pornography, and malignant stereotypes, 2021.

\bibitem[Birhane et~al.(2024)Birhane, Han, Boddeti, Luccioni, et~al.]{birhane2024into}
Abeba Birhane, Sanghyun Han, Vishnu Boddeti, Sasha Luccioni, et~al.
\newblock Into the {LAION}’s den: Investigating hate in multimodal datasets.
\newblock In \emph{NeurIPS}, 2024.

\bibitem[Carion et~al.(2020)Carion, Massa, Synnaeve, Usunier, Kirillov, and Zagoruyko]{carion2020end}
Nicolas Carion, Francisco Massa, Gabriel Synnaeve, Nicolas Usunier, Alexander Kirillov, and Sergey Zagoruyko.
\newblock End-to-end object detection with {T}ransformers.
\newblock In \emph{ECCV}, 2020.

\bibitem[Caron et~al.(2021)Caron, Touvron, Misra, J{\'e}gou, Mairal, Bojanowski, and Joulin]{caron2021emerging}
Mathilde Caron, Hugo Touvron, Ishan Misra, Herv{\'e} J{\'e}gou, Julien Mairal, Piotr Bojanowski, and Armand Joulin.
\newblock Emerging properties in self-supervised {V}ision {T}ransformers.
\newblock In \emph{ICCV}, 2021.

\bibitem[Chambers et~al.(2010)Chambers, Raniwala, Perry, Adams, Henry, Bradshaw, and Weizenbaum]{chambers2010flumejava}
Craig Chambers, Ashish Raniwala, Frances Perry, Stephen Adams, Robert~R Henry, Robert Bradshaw, and Nathan Weizenbaum.
\newblock Flume{J}ava: easy, efficient data-parallel pipelines.
\newblock \emph{ACM SIGPLAN Notices}, 2010.

\bibitem[Chen et~al.(2024)Chen, Hirota, Otani, Garcia, and Nakashima]{chen2024would}
Tianwei Chen, Yusuke Hirota, Mayu Otani, Noa Garcia, and Yuta Nakashima.
\newblock Would deep generative models amplify bias in future models?
\newblock In \emph{CVPR}, 2024.

\bibitem[Chen et~al.(2020)Chen, Li, Yu, El~Kholy, Ahmed, Gan, Cheng, and Liu]{10.1007/978-3-030-58577-8_7}
Yen-Chun Chen, Linjie Li, Licheng Yu, Ahmed El~Kholy, Faisal Ahmed, Zhe Gan, Yu Cheng, and Jingjing Liu.
\newblock {UNITER}: {UN}iversal {I}mage-{TE}xt {R}epresentation learning.
\newblock In \emph{ECCV}, 2020.

\bibitem[Cherti et~al.(2023)Cherti, Beaumont, Wightman, Wortsman, Ilharco, Gordon, Schuhmann, Schmidt, and Jitsev]{cherti2023reproducible}
Mehdi Cherti, Romain Beaumont, Ross Wightman, Mitchell Wortsman, Gabriel Ilharco, Cade Gordon, Christoph Schuhmann, Ludwig Schmidt, and Jenia Jitsev.
\newblock Reproducible scaling laws for contrastive language-image learning.
\newblock In \emph{CVPR}, 2023.

\bibitem[Deng et~al.(2024)Deng, Zhao, Tang, Gerstein, and Cohan]{deng2024investigating}
Chunyuan Deng, Yilun Zhao, Xiangru Tang, Mark Gerstein, and Arman Cohan.
\newblock Investigating data contamination in modern benchmarks for large language models.
\newblock In \emph{NAACL}, 2024.

\bibitem[Deng et~al.(2009)Deng, Dong, Socher, Li, Li, and Fei-Fei]{deng2009imagenet}
Jia Deng, Wei Dong, Richard Socher, Li-Jia Li, Kai Li, and Li Fei-Fei.
\newblock Image{N}et: A large-scale hierarchical image database.
\newblock In \emph{CVPR}, 2009.

\bibitem[Dodge et~al.(2021)Dodge, Sap, Marasovi{\'c}, Agnew, Ilharco, Groeneveld, Mitchell, and Gardner]{dodge2021documenting}
Jesse Dodge, Maarten Sap, Ana Marasovi{\'c}, William Agnew, Gabriel Ilharco, Dirk Groeneveld, Margaret Mitchell, and Matt Gardner.
\newblock Documenting large webtext corpora: A case study on the colossal clean crawled corpus.
\newblock In \emph{EMNLP}, 2021.

\bibitem[Douze et~al.(2025)Douze, Guzhva, Deng, Johnson, Szilvasy, Mazaré, Lomeli, Hosseini, and Jégou]{douze2024faiss}
Matthijs Douze, Alexandr Guzhva, Chengqi Deng, Jeff Johnson, Gergely Szilvasy, Pierre-Emmanuel Mazaré, Maria Lomeli, Lucas Hosseini, and Hervé Jégou.
\newblock The {F}aiss library, 2025.

\bibitem[Elazar et~al.(2024)Elazar, Bhagia, Magnusson, Ravichander, Schwenk, Suhr, Walsh, Groeneveld, Soldaini, Singh, et~al.]{elazar2024what}
Yanai Elazar, Akshita Bhagia, Ian~Helgi Magnusson, Abhilasha Ravichander, Dustin Schwenk, Alane Suhr, Evan~Pete Walsh, Dirk Groeneveld, Luca Soldaini, Sameer Singh, et~al.
\newblock What's in my big data?
\newblock In \emph{ICLR}, 2024.

\bibitem[Gadre et~al.(2024)Gadre, Ilharco, Fang, Hayase, Smyrnis, Nguyen, Marten, Wortsman, Ghosh, Zhang, et~al.]{gadre2024datacomp}
Samir~Yitzhak Gadre, Gabriel Ilharco, Alex Fang, Jonathan Hayase, Georgios Smyrnis, Thao Nguyen, Ryan Marten, Mitchell Wortsman, Dhruba Ghosh, Jieyu Zhang, et~al.
\newblock Data{C}omp: In search of the next generation of multimodal datasets.
\newblock In \emph{NeurIPS}, 2024.

\bibitem[Garcia et~al.(2023)Garcia, Hirota, Wu, and Nakashima]{garcia2023uncurated}
Noa Garcia, Yusuke Hirota, Yankun Wu, and Yuta Nakashima.
\newblock Uncurated image-text datasets: Shedding light on demographic bias.
\newblock In \emph{CVPR}, 2023.

\bibitem[Girshick(2015)]{girshick2015fast}
R Girshick.
\newblock Fast {R-CNN}.
\newblock In \emph{ICCV}, 2015.

\bibitem[Golchin and Surdeanu(2024)]{golchintime}
Shahriar Golchin and Mihai Surdeanu.
\newblock Time travel in {LLM}s: Tracing data contamination in large language models.
\newblock In \emph{ICLR}, 2024.

\bibitem[He et~al.(2016)He, Zhang, Ren, and Sun]{he2016deep}
Kaiming He, Xiangyu Zhang, Shaoqing Ren, and Jian Sun.
\newblock Deep residual learning for image recognition.
\newblock In \emph{CVPR}, 2016.

\bibitem[Jiang et~al.(2023)Jiang, Sablayrolles, Mensch, Bamford, Chaplot, Casas, Bressand, Lengyel, Lample, Saulnier, et~al.]{jiang2023mistral}
Albert~Q Jiang, Alexandre Sablayrolles, Arthur Mensch, Chris Bamford, Devendra~Singh Chaplot, Diego de~las Casas, Florian Bressand, Gianna Lengyel, Guillaume Lample, Lucile Saulnier, et~al.
\newblock Mistral 7{B}, 2023.

\bibitem[Johnson et~al.(2017)Johnson, Hariharan, Van Der~Maaten, Fei-Fei, Lawrence~Zitnick, and Girshick]{johnson2017clevr}
Justin Johnson, Bharath Hariharan, Laurens Van Der~Maaten, Li Fei-Fei, C Lawrence~Zitnick, and Ross Girshick.
\newblock {CLEVR}: A diagnostic dataset for compositional language and elementary visual reasoning.
\newblock In \emph{CVPR}, 2017.

\bibitem[Karpathy and Fei-Fei(2015)]{karpathy2015deep}
Andrej Karpathy and Li Fei-Fei.
\newblock Deep visual-semantic alignments for generating image descriptions.
\newblock In \emph{CVPR}, 2015.

\bibitem[Kiela et~al.(2020)Kiela, Firooz, Mohan, Goswami, Singh, Ringshia, and Testuggine]{kiela2020hateful}
Douwe Kiela, Hamed Firooz, Aravind Mohan, Vedanuj Goswami, Amanpreet Singh, Pratik Ringshia, and Davide Testuggine.
\newblock The {H}ateful {M}emes {C}hallenge: Detecting hate speech in multimodal memes.
\newblock In \emph{NeurIPS}, 2020.

\bibitem[Kuznetsova et~al.(2020)Kuznetsova, Rom, Alldrin, Uijlings, Krasin, Pont-Tuset, Kamali, Popov, Malloci, Kolesnikov, et~al.]{kuznetsova2020open}
Alina Kuznetsova, Hassan Rom, Neil Alldrin, Jasper Uijlings, Ivan Krasin, Jordi Pont-Tuset, Shahab Kamali, Stefan Popov, Matteo Malloci, Alexander Kolesnikov, et~al.
\newblock The {O}pen {I}mages {D}ataset {V}4: Unified image classification, object detection, and visual relationship detection at scale.
\newblock \emph{IJCV}, 2020.

\bibitem[LeCun et~al.(1998)LeCun, Bottou, Bengio, and Haffner]{lecun1998gradient}
Yann LeCun, L{\'e}on Bottou, Yoshua Bengio, and Patrick Haffner.
\newblock Gradient-based learning applied to document recognition.
\newblock \emph{Proc. IEEE}, 1998.

\bibitem[Li et~al.(2024)Li, Wong, Zhang, Usuyama, Liu, Yang, Naumann, Poon, and Gao]{li2024llava}
Chunyuan Li, Cliff Wong, Sheng Zhang, Naoto Usuyama, Haotian Liu, Jianwei Yang, Tristan Naumann, Hoifung Poon, and Jianfeng Gao.
\newblock {LL}a{VA-M}ed: Training a large language-and-vision assistant for biomedicine in one day.
\newblock In \emph{NeurIPS}, 2024.

\bibitem[Lin et~al.(2014)Lin, Maire, Belongie, Hays, Perona, Ramanan, Doll{\'a}r, and Zitnick]{lin2014microsoft}
Tsung-Yi Lin, Michael Maire, Serge Belongie, James Hays, Pietro Perona, Deva Ramanan, Piotr Doll{\'a}r, and C~Lawrence Zitnick.
\newblock Microsoft {COCO}: Common objects in context.
\newblock In \emph{ECCV}, 2014.

\bibitem[Liu et~al.(2021)Liu, Lin, Cao, Hu, Wei, Zhang, Lin, and Guo]{liu2021swin}
Ze Liu, Yutong Lin, Yue Cao, Han Hu, Yixuan Wei, Zheng Zhang, Stephen Lin, and Baining Guo.
\newblock Swin {T}ransformer: Hierarchical {V}ision {T}ransformer using shifted windows.
\newblock In \emph{ICCV}, 2021.

\bibitem[Magar and Schwartz(2022)]{magar2022data}
Inbal Magar and Roy Schwartz.
\newblock Data contamination: From memorization to exploitation.
\newblock In \emph{ACL}, 2022.

\bibitem[Mahajan et~al.(2018)Mahajan, Girshick, Ramanathan, He, Paluri, Li, Bharambe, and Van Der~Maaten]{mahajan2018exploring}
Dhruv Mahajan, Ross Girshick, Vignesh Ramanathan, Kaiming He, Manohar Paluri, Yixuan Li, Ashwin Bharambe, and Laurens Van Der~Maaten.
\newblock Exploring the limits of weakly supervised pretraining.
\newblock In \emph{ECCV}, 2018.

\bibitem[Meister et~al.(2023)Meister, Zhao, Wang, Ramaswamy, Fong, and Russakovsky]{meister2023gender}
Nicole Meister, Dora Zhao, Angelina Wang, Vikram~V Ramaswamy, Ruth Fong, and Olga Russakovsky.
\newblock Gender artifacts in visual datasets.
\newblock In \emph{ICCV}, 2023.

\bibitem[Oquab et~al.(2024)Oquab, Darcet, Moutakanni, Vo, Szafraniec, Khalidov, Fernandez, Haziza, Massa, El-Nouby, et~al.]{oquab2023dinov2}
Maxime Oquab, Timoth{\'e}e Darcet, Th{\'e}o Moutakanni, Huy Vo, Marc Szafraniec, Vasil Khalidov, Pierre Fernandez, Daniel Haziza, Francisco Massa, Alaaeldin El-Nouby, et~al.
\newblock {DINO}v2: Learning robust visual features without supervision.
\newblock In \emph{ICLR}, 2024.

\bibitem[Oren et~al.(2024)Oren, Meister, Chatterji, Ladhak, and Hashimoto]{oren2024proving}
Yonatan Oren, Nicole Meister, Niladri~S Chatterji, Faisal Ladhak, and Tatsunori Hashimoto.
\newblock Proving test set contamination in black-box language models.
\newblock In \emph{ICLR}, 2024.

\bibitem[Piktus et~al.(2023)Piktus, Akiki, Villegas, Lauren{\c{c}}on, Dupont, Luccioni, Jernite, and Rogers]{piktus-etal-2023-roots}
Aleksandra Piktus, Christopher Akiki, Paulo Villegas, Hugo Lauren{\c{c}}on, G{\'e}rard Dupont, Sasha Luccioni, Yacine Jernite, and Anna Rogers.
\newblock The {ROOTS} search tool: Data transparency for {LLM}s.
\newblock In \emph{ACL}, 2023.

\bibitem[Radford et~al.(2021)Radford, Kim, Hallacy, Ramesh, Goh, Agarwal, Sastry, Askell, Mishkin, Clark, et~al.]{radford2021learning}
Alec Radford, Jong~Wook Kim, Chris Hallacy, Aditya Ramesh, Gabriel Goh, Sandhini Agarwal, Girish Sastry, Amanda Askell, Pamela Mishkin, Jack Clark, et~al.
\newblock Learning transferable visual models from natural language supervision.
\newblock In \emph{ICML}, 2021.

\bibitem[Ravaut et~al.(2025)Ravaut, Ding, Jiao, Chen, Li, Zhao, Qin, Xiong, and Joty]{ravaut2024much}
Mathieu Ravaut, Bosheng Ding, Fangkai Jiao, Hailin Chen, Xingxuan Li, Ruochen Zhao, Chengwei Qin, Caiming Xiong, and Shafiq Joty.
\newblock How much are {LLM}s contaminated? a comprehensive survey and the llmsanitize library, 2025.

\bibitem[Ren et~al.(2016)Ren, He, Girshick, and Sun]{ren2016faster}
Shaoqing Ren, Kaiming He, Ross Girshick, and Jian Sun.
\newblock Faster {R}-{CNN}: Towards real-time object detection with region proposal networks.
\newblock \emph{PAMI}, 2016.

\bibitem[Roberts et~al.(2023)Roberts, Thakur, Herlihy, White, and Dooley]{roberts2023cutoff}
Manley Roberts, Himanshu Thakur, Christine Herlihy, Colin White, and Samuel Dooley.
\newblock To the cutoff... and beyond? a longitudinal perspective on {LLM} data contamination.
\newblock In \emph{ICLR}, 2023.

\bibitem[Rombach et~al.(2022)Rombach, Blattmann, Lorenz, Esser, and Ommer]{rombach2021highresolution}
Robin Rombach, Andreas Blattmann, Dominik Lorenz, Patrick Esser, and Björn Ommer.
\newblock High-resolution image synthesis with latent diffusion models.
\newblock In \emph{CVPR}, 2022.

\bibitem[Sainz et~al.(2023)Sainz, Campos, Garc{\'\i}a-Ferrero, Etxaniz, de~Lacalle, and Agirre]{sainz-etal-2023-nlp}
Oscar Sainz, Jon Campos, Iker Garc{\'\i}a-Ferrero, Julen Etxaniz, Oier~Lopez de Lacalle, and Eneko Agirre.
\newblock {NLP} evaluation in trouble: On the need to measure {LLM} data contamination for each benchmark.
\newblock In \emph{EMNLP Findings}, 2023.

\bibitem[Schuhmann et~al.(2021)Schuhmann, Kaczmarczyk, Komatsuzaki, Katta, Vencu, Beaumont, Jitsev, Coombes, and Mullis]{schuhmann2021laion}
Christoph Schuhmann, Robert Kaczmarczyk, Aran Komatsuzaki, Aarush Katta, Richard Vencu, Romain Beaumont, Jenia Jitsev, Theo Coombes, and Clayton Mullis.
\newblock {LAION}-400{M}: Open dataset of {CLIP}-filtered 400 million image-text pairs.
\newblock In \emph{NeurIPSW}, 2021.

\bibitem[Schuhmann et~al.(2022)Schuhmann, Beaumont, Vencu, Gordon, Wightman, Cherti, Coombes, Katta, Mullis, Wortsman, et~al.]{schuhmann2022laion}
Christoph Schuhmann, Romain Beaumont, Richard Vencu, Cade Gordon, Ross Wightman, Mehdi Cherti, Theo Coombes, Aarush Katta, Clayton Mullis, Mitchell Wortsman, et~al.
\newblock {LAION}-5{B}: An open large-scale dataset for training next generation image-text models.
\newblock In \emph{NeurIPS}, 2022.

\bibitem[Sharma et~al.(2018)Sharma, Ding, Goodman, and Soricut]{sharma2018conceptual}
Piyush Sharma, Nan Ding, Sebastian Goodman, and Radu Soricut.
\newblock Conceptual {C}aptions: A cleaned, hypernymed, image alt-text dataset for automatic image captioning.
\newblock In \emph{ACL}, 2018.

\bibitem[Shi et~al.(2024)Shi, Ajith, Xia, Huang, Liu, Blevins, Chen, and Zettlemoyer]{shi2024detecting}
Weijia Shi, Anirudh Ajith, Mengzhou Xia, Yangsibo Huang, Daogao Liu, Terra Blevins, Danqi Chen, and Luke Zettlemoyer.
\newblock Detecting pretraining data from large language models.
\newblock In \emph{ICLR}, 2024.

\bibitem[Sidorov et~al.(2020)Sidorov, Hu, Rohrbach, and Singh]{sidorov2020textcaps}
Oleksii Sidorov, Ronghang Hu, Marcus Rohrbach, and Amanpreet Singh.
\newblock Text{C}aps: a dataset for image captioning with reading comprehension.
\newblock In \emph{ECCV}, 2020.

\bibitem[Su et~al.(2020)Su, Zhu, Cao, Li, Lu, Wei, and Dai]{Su2020VL-BERT}
Weijie Su, Xizhou Zhu, Yue Cao, Bin Li, Lewei Lu, Furu Wei, and Jifeng Dai.
\newblock {VL}-{BERT}: Pre-training of generic visual-linguistic representations.
\newblock In \emph{ICLR}, 2020.

\bibitem[Tolias et~al.(2016)Tolias, Sicre, and J{\'e}gou]{tolias2016particular}
Giorgos Tolias, Ronan Sicre, and Herv{\'e} J{\'e}gou.
\newblock Particular object retrieval with integral max-pooling of {CNN} activations.
\newblock In \emph{ICLR}, 2016.

\bibitem[Touvron et~al.(2023)Touvron, Martin, Stone, Albert, Almahairi, Babaei, Bashlykov, Batra, Bhargava, Bhosale, et~al.]{touvron2023llama}
Hugo Touvron, Louis Martin, Kevin Stone, Peter Albert, Amjad Almahairi, Yasmine Babaei, Nikolay Bashlykov, Soumya Batra, Prajjwal Bhargava, Shruti Bhosale, et~al.
\newblock Llama 2: Open foundation and fine-tuned chat models, 2023.

\bibitem[Vinyals et~al.(2015)Vinyals, Toshev, Bengio, and Erhan]{vinyals2015show}
Oriol Vinyals, Alexander Toshev, Samy Bengio, and Dumitru Erhan.
\newblock Show and tell: A neural image caption generator.
\newblock In \emph{CVPR}, 2015.

\bibitem[Wah et~al.(2011)Wah, Branson, Welinder, Perona, and Belongie]{wah2011caltech}
Catherine Wah, Steve Branson, Peter Welinder, Pietro Perona, and Serge Belongie.
\newblock The {C}altech-{UCSD} {B}irds-200-2011 {D}ataset.
\newblock Technical report, California Institute of Technology, 2011.

\bibitem[Wang et~al.(2025)Wang, Yeh, and Mark~Liao]{wang2025yolov9}
Chien-Yao Wang, I-Hau Yeh, and Hong-Yuan Mark~Liao.
\newblock {YOLO}v9: Learning what you want to learn using programmable gradient information.
\newblock In \emph{ECCV}, 2025.

\bibitem[Young et~al.(2014)Young, Lai, Hodosh, and Hockenmaier]{young-etal-2014-image}
Peter Young, Alice Lai, Micah Hodosh, and Julia Hockenmaier.
\newblock From image descriptions to visual denotations: New similarity metrics for semantic inference over event descriptions.
\newblock \emph{TACL}, 2014.

\bibitem[Zauner(2010)]{zauner2010implementation}
Christoph Zauner.
\newblock Implementation and benchmarking of perceptual image hash functions.
\newblock Master's thesis, Upper Austria University of Applied Sciences, 2010.

\bibitem[Zhai et~al.(2022)Zhai, Wang, Mustafa, Steiner, Keysers, Kolesnikov, and Beyer]{zhai2022lit}
Xiaohua Zhai, Xiao Wang, Basil Mustafa, Andreas Steiner, Daniel Keysers, Alexander Kolesnikov, and Lucas Beyer.
\newblock Li{T}: Zero-shot transfer with locked-image text tuning.
\newblock In \emph{CVPR}, 2022.

\bibitem[Zhou et~al.(2017)Zhou, Lapedriza, Khosla, Oliva, and Torralba]{zhou2017places}
Bolei Zhou, Agata Lapedriza, Aditya Khosla, Aude Oliva, and Antonio Torralba.
\newblock Places: A 10 million image database for scene recognition.
\newblock \emph{PAMI}, 2017.

\end{thebibliography}
